\documentclass{article}
\pdfoutput=1

\usepackage[preprint]{nips_2018}

\usepackage[utf8]{inputenc} 
\usepackage[T1]{fontenc}    
\usepackage{hyperref}       
\usepackage{url}            
\usepackage{booktabs}       
\usepackage{amsfonts}       
\usepackage{nicefrac}       
\usepackage{microtype}      
\usepackage{empheq}
\usepackage{mathtools}
\usepackage{graphics} 
\usepackage{graphicx} 
\usepackage{times} 
\usepackage{amsmath} 
\usepackage{amssymb}  
\usepackage{color}
\usepackage{subfig}
\usepackage{epstopdf}
\usepackage{wrapfig}
\usepackage{adjustbox}
\usepackage{booktabs}
\usepackage{multirow}
\usepackage{balance}
\usepackage{commath}
\usepackage{amsmath}
\usepackage{xcolor}
\usepackage{gensymb}
\usepackage[nobreak=true]{mdframed}
\usepackage{kantlipsum}
\usepackage{afterpage}
\usepackage{algorithm}
\usepackage[noend]{algpseudocode}
\usepackage{newfloat}
\DeclareFloatingEnvironment[fileext=frm,placement={!ht},name=Frame]{myfloat}
\makeatletter
\newenvironment{AlgoBox}[2]{ 
\begin{myfloat}[tb!]
\protected@edef\@currentlabelname{#1}
\protected@edef\@currentlabel{#2}
\begin{mdframed}[
innerlinewidth=0.5pt,
innerleftmargin=10pt,
innerrightmargin=10pt,
innertopmargin = 10pt,
innerbottommargin=10pt,
skipabove=\dimexpr\topsep+\ht\strutbox\relax,
roundcorner=5pt,
frametitle={#1},
frametitlerule=true,
frametitlerulewidth=1pt]
}{
\end{mdframed}
\end{myfloat}
}
\makeatother

\title{Pull Message Passing for \\ Nonparametric  Belief Propagation}

\author{
	Karthik Desingh
    \And
    Anthony Opipari
    \And 
    Odest Chadwicke Jenkins \\
    \AND
    \\ Laboratory for Perception, RObotics, and Grounded REasoning SystemS \\
    Department of Electrical Engineering and Computer Science \\
    Robotics Institute \\
    University of Michigan \\
    Ann Arbor, MI, USA  48105 \\
    \texttt{kdesingh|topipari|ocj@umich.edu}
}

\begin{document}

\maketitle

\begin{abstract}

We present a ``pull'' approach to approximate products of Gaussian mixtures within message updates for Nonparametric Belief Propagation (NBP) inference.  Existing NBP methods often represent messages between continuous-valued latent variables as Gaussian mixture models. To avoid computational intractability in loopy graphs, NBP necessitates an approximation of the product of such mixtures.  
Sampling-based product approximations have shown effectiveness for NBP inference.  However, such approximations used within the traditional ``push'' message update procedures quickly become computationally prohibitive for multi-modal distributions over high-dimensional variables. In contrast, we propose a ``pull'' method, as the Pull Message Passing for Nonparametric Belief propagation (PMPNBP) algorithm, and demonstrate its viability for efficient inference. We report results using an experiment from an existing NBP method, PAMPAS, for inferring the pose of an articulated structure in clutter.  Results from this illustrative problem found PMPNBP has a greater ability to efficiently scale the number of components in its mixtures and, consequently, improve inference accuracy.

\end{abstract}

\section{Introduction}
\label{introduction}
We present the Pull Message Passing Nonparametric Belief Propagation (PMPNBP) algorithm as a ``pull'' approach to approximating message productions for loopy belief propagation in a Markov Random Field model.  Building on existing methods for Nonparametric Belief Propagation (NBP), PMPNBP aims to perform inference of continuous, high-dimensional, and multi-modal random variables.  Propagation of belief in NBP involves updating messages that inform the belief of one random variable based on the belief of another variable.  This message update typically requires taking the product of $d$  mixture models, each with $M$ Gaussian components.  In exact form, the distribution resulting from this product will be comprised of $M^d$ Gaussian components.  Consequently, the number of components needed to represent messages will grow towards intractability with respect to the number of update iterations.  

Existing methods for NBP commonly address this intractability through a sampling-based approximation within a ``push'' message update procedure.  This push procedure updates a message by first approximating the mixture product, often by Gibbs sampling, and then propagating this product to the message update.  While effective for accurate inference, Gibbs sampling is computationally costly and prohibitive for many applications with time-critical demands and bounded computational resources, such as in robotics. More specifically, push updating in this manner suffers from two critical issues: 1) the computational cost incurred for iterative sampling of the approximated product, and 2) the limited number of mixture components that can be asymptotically accommodated. 


Consider the problem of robot perception in cluttered scenes~\cite{desingh2016physically,narayanan2016discriminatively,papazov2012rigid,sui2017goal}.
Such scene perception requires inference over continuous-valued pose spaces for a varying number of objects.  Inference in these continuous spaces must also contend with high dimensionality, scaling with the number of objects, and multi-modal distributions, due to partial ambiguous observations.  A vast body of existing literature has explored methods to address this type of inference problem \cite{sigal2004tracking,sudderth2004visual,vondrak2013dynamical}.  Among these methods, we focus our attention on algorithms for inference by belief propagation in loopy probabilistic graphical models. In particular, Nonparametric Belief Propagation ~\cite{isard2003pampas,sudderth2003nonparametric} has demonstrated considerable potential to address the challenges of inference for continuous, high-dimensional, and multi-modal random variables. However, direct application of these methods remains a substantial computational investment and intellectual challenge. 

In this paper, we address the computational challenges of existing NBP methods and provide a more efficient ``pull'' message passing approach through the PMPNBP algorithm.  The key idea of pull message updating is to evaluate samples taken from the belief of the receiving node with respect to the densities informing the sending node.  The mixture product approximation can then be performed individually per sample, and then later normalized into a distribution.  This pull updating avoids the computational pitfalls of push updating of message distributions, which requires exponential growth in the number of components or expensive iterative methods.
We demonstrate the accuracy and efficiency of inference by PMPNBP with respect to PAMPAS~\cite{isard2003pampas}, a pioneering method for NBP.  These results focus on an experiment for finding an articulated 2D pattern, reconstructed from the description of PAMPAS.  These results indicate PMPNBP enables both faster convergence to an appropriate inference and greater scaling of message mixture components for improved accuracy.

\section{Related work}
\label{related_work}
Probabilistic graphical models, such as the Markov Random Field (MRF), are widely used in computational perception to model problems involving inference of random variables under considerable uncertainty.  Many algorithms have been proposed to compute the joint probability of graphical models in these cases. Belief propagation algorithms are a category of algorithms that are guaranteed to converge on tree-structured graphs. For graph structures with loops, Loopy Belief Propagation (LBP)~\cite{murphy1999loopy} is empirically proven to perform for discrete variables. Recently, Chua et al.~\cite{chua2016scene} proposed a belief propagation over factor graphs to generate scenes satisfying the scene grammars. The problem becomes much more challenging when the latent variables take continuous values. Sudderth et al. ~\cite{sudderth2003nonparametric} and Isard et al.~\cite{isard2003pampas} introduced methods for nonparametric belief propagation to address such continuous-valued cases.  Both of these approaches approximate a continuous-valued function as a mixture of weighted Gaussians and use local Gibbs sampling to approximate the product of mixtures. This approach to message passing has been effectively used in applications such as human pose estimation~\cite{sigal2004tracking} and hand tracking~\cite{sudderth2004visual} by modeling the graph as a tree structured particle network. In order to viably pursue NBP for robotic problems, such as scene perception, the computational efficiency of NBP methods needs to be revisited. 


Some recent works address the computational efficiency of Nonparametric Belief Propagation.  Similar in spirit to PMPNBB, Ihler et. al~\cite{ihler2009particle} describe a conceptual theory of particle belief propagation, where a target node's samples are used to generate a message going from source to target. This work emphasizes the advantages of using large number of particles to represent incoming messages, along with theoretical  analysis.  This work uses an expensive iterative Markov Chain Monte Carlo sampling step, mimicking the Gibbs sampling step in NBP~\cite{isard2003pampas,sudderth2003nonparametric}.  PMPNBP is able to avoid this cost through a resampling step. 

Kernel based methods have been proposed to improve the efficiency of NBP. Song et. al~\cite{song2011kernel} propose a kernel belief propagation method. Messages in this work are represented as functions in a Reproducing Kernel Hilbert space (RKHS) and message updates are linear operations in RKHS. Results presented in this work claim to be more accurate and faster than NBP with Gibbs sampling ~\cite{isard2003pampas,sudderth2003nonparametric} and particle belief propagation~\cite{ihler2009particle} over applications such as image denoising, depth prediction, and angle prediction in protein folding problem. We consider comparisons with kernel-based approximators as a direction for future work.
Han et. al~\cite{han2006efficientnb} introduces mode propagation to approximate the slow sampling based products in NBP~\cite{isard2003pampas,sudderth2003nonparametric} with a few mode propagation and kernel fitting steps. However, their approach is limited to non-occluded observations. Our proposed algorithm PMPNBP handles occlusions with convergence characteristics comparable to PAMPAS~\cite{isard2003pampas}.



\section{Nonparametric Belief Propagation}
\label{nbp_overview}
Let $G=(V, E)$ be an undirected graph with nodes $V$ and edges $E$. The nodes in $V$ are each random variables that have dependencies with each other in the graph $G$ through edges $E$. If $G$ is a Markov Random Field (MRF), then it has two types of variables $X$ and $Y$, denoting the collection of hidden and observed variables, respectively.  Each variable is considered to take assignments of continuous-valued vectors. The joint probability of the graph $G$, considering only second order cliques, is given as
\begin{equation}\label{eq:1}
p( X, Y ) = \frac{1}{Z}\prod_{(s, t)\in E} \psi_{s, t}(X_s, X_t) \prod_{s \in V} \phi_s(X_s, Y_s)
\end{equation}
where $\psi_{s, t}(X_s, X_t)$ is the pairwise potential between nodes $X_s \in \mathbb{R}^d$ and $X_t \in \mathbb{R}^b$ \footnote{Note, dimensionality remains the same, $d=b$, in the case of estimating 6 degree-of-freedom object pose}, 
$\phi_s(X_s, Y_s)$ is the unary potential between the hidden node $X_s$ and observed node $Y_s \in \mathbb{R}^q$, and $Z$ is a normalizing factor. The problem is to infer belief over possible states assigned to the hidden variables $X$ such that the joint probability is maximized. This inference is generally performed by passing messages between hidden variables $X$ until convergence of their belief distributions over several iterations. 

A message is denoted as $m_{t\rightarrow s}$ directed from node $t$ to node $s$ if there is an edge between the nodes in the graph $G$. The message represents the distribution of what node $t$ thinks node $s$ should take in terms of the hidden variable $X_s$. Typically, if $X_s$ is in the continuous domain, then $m_{t\rightarrow s}(X_s)$ is represented as a Gaussian mixture to approximate the real distribution: 
\begin{equation}\label{eq:2}
m_{t\rightarrow s}(X_s) = \sum_{i=1}^M w_{ts}^{(i)} \mathcal{N}(X_s; \mu_{ts}^{(i)}, \Lambda_{ts}^{(i)})
\end{equation}
where $\sum_{i=1}^M w_{ts}^{(i)} = 1$, $M$ is the number of Gaussian components, $w_{ts}^{(i)}$ is the weight associated with the $i^{th}$ component, $\mu_{ts}^{(i)}$ and $\Lambda_{ts}^{(i)}$ are the mean and covariance of the $i^{th}$ component, respectively. We use the terms components, particles and samples interchangeably in this paper. Hence, a message can be expressed as $M$ triplets:

\begin{equation}\label{eq:3}
m_{t\rightarrow s} = \{(w_{ts}^{(i)}, \mu_{ts}^{(i)}, \Lambda_{ts}^{(i)}): 1\leq i \leq M\}
\end{equation}

Assuming the graph has tree or loopy structure, computing these message updates is nontrivial computationally.
A message update in a continuous domain at an iteration $n$ from a node $t \rightarrow s$ is given by
\begin{equation}\label{eq:4}
m_{t\rightarrow s}^n(X_s) \leftarrow \int_{X_t \in \mathbb{R}^b} \bigg(\psi_{st}(X_s, X_t)\phi_t(X_t, Y_t) \prod_{u \in \rho(t)\setminus s} m_{u\rightarrow t}^{n-1}(X_t)\bigg)dX_t 
\end{equation}
where $\rho(t)$ is a set of neighbor nodes of $t$. The marginal belief over each hidden node at iteration $n$ is given by
\begin{equation}\label{eq:5}
\begin{aligned}
bel_s^n(X_s) \propto \phi_s(X_s, Y_s) \prod_{t \in \rho(s)} m_{t\rightarrow s}^n(X_s) \\
bel_s^n = \{(w_s^{(i)}, \mu_s^{(i)}, \Lambda_s^{(i)}): 1\leq i \leq T\}
\end{aligned}
\end{equation}
where $T$ is the number of components used to represent the belief.
NBP~\cite{sudderth2003nonparametric} provides a Gibbs sampling approach to compute an approximation of the product $\prod_{u \in \rho(t)\setminus s} m_{u\rightarrow t}^{n-1}(X_t)$. 
Assuming that $\phi_t(X_t, Y_t)$ is pointwise computable, a ``pre-message'' \cite{ihler2009particle} is defined as
\begin{equation}\label{eq:6}
M_{t\rightarrow s}^{n-1}(X_t)=\phi_t(X_t, Y_t)\prod_{u \in \rho(t)\setminus s} m_{u\rightarrow t}^{n-1}(X_t)
\end{equation}
which can be computed in the Gibbs sampling procedure. This reduces Equation~\ref{eq:4} to
\begin{equation}\label{eq:7}
m_{t\rightarrow s}^n(X_s) \leftarrow \int_{X_t \in \mathbb{R}^b}  \bigg(\psi_{st}(X_s, X_t)M_{t\rightarrow s}^{n-1}(X_t)\bigg)dX_t
\end{equation}

\begin{AlgoBox}{\small Algorithm - Message update}{Algorithm - Message update}\label{alg:message_update}
\small
Given input messages $m_{u\rightarrow t}^{n-1}(X_t) = \{(\mu_{ut}^{(i)}, w_{ut}^{(i)})\}_{i=1}^{M}$ for each $u \in \rho(t)\setminus s$, and methods to compute functions $\psi_{ts}(X_t, X_s)$ and $\phi_t(X_t, Y_t)$ point-wise, the algorithm computes $m_{t\rightarrow s}^{n}(X_s)=\{(\mu_{ts}^{(i)}, w_{ts}^{(i)})\}_{i=1}^M$
\begin{enumerate}
    \item[1.] Draw $M$ independent samples $\{\mu_{ts}^{(i)}\}_{i=1}^M$ from $bel_s^{n-1}(X_s)$.
    \begin{enumerate}
        \item [(a)] If $n=1$ the $bel_s^{0}(X_s)$ is a uniform distribution or informed by a prior distribution.
        \item [(b)] If $n>1$ the $bel_s^{n-1}(X_s)$ is a belief computed at $(n-1)^{th}$ iteration using importance sampling.  
    \end{enumerate}
    \item[2] For each $\{\mu_{ts}^{(i)}\}_{i=1}^M$, compute $w_{ts}^{(i)}$
    \begin{enumerate}
    \item[a] Sample $\hat{X}_t^{(i)} \sim \psi_{ts}(X_t, X_s=\mu_{ts}^{(i)})$ 
    \item[b] Unary weight $w_{unary}^{(i)}$ is computed using $\phi_t(X_t=\hat{X}_t^{(i)}, Y_t)$.
    \item[c] Neighboring weight $w_{neigh}^{(i)}$ is computed using $m_{u\rightarrow t}^{n-1}$.
    \begin{enumerate}
        \item[(i)] For each $u \in \rho(t) \setminus s$ compute $W_u^{(i)}=\sum_{j=1}^Mw_{ut}^{(j)}w_{u}^{(ij)}$ where \\ $w_{u}^{(ij)} = \psi_{ts}(X_s=\mu_{ts}^{(i)}, X_t=\mu_{ut}^{(j)})$.
        
        \item[(ii)] Each neighboring weight is computed by $w_{neigh}^{(i)}=\prod_{u \in \rho(t) \setminus s}W_{u}^{(i)}$
    \end{enumerate}
    \item[d] The final weights are computed as $w_{ts}^{(i)}=w_{neigh}^{(i)} \times w_{unary}^{(i)}$. 
    \end{enumerate}
    \item[3] The weights $\{w_{ts}^{(i)}\}_{i=1}^M$ are associated with the samples $\{\mu_{ts}^{(i)}\}_{i=1}^M$ to represent $m_{t\rightarrow s}^n(X_s)$.
\end{enumerate}
\end{AlgoBox}
\begin{AlgoBox}{\small Algorithm - Belief update}{Algorithm - Belief update}\label{alg:belief_update}
Given incoming messages $m_{t\rightarrow s}^n(X_t) = \{( w_{ts}^{(i)},\mu_{ts}^{(i)})\}_{i=1}^{M}$ for each $t \in \rho(s)$, and methods to compute functions $\phi_s(x_s, y_s)$ point-wise, the algorithm computes $bel_s^n(X_s) \propto \phi_s(X_s, Y_s) \prod_{t \in \rho(s)} m_{t\rightarrow s}^n(X_s) = \{( w_{s}^{(i)},\mu_{s}^{(i)})\}_{i=1}^{T}$
\begin{enumerate}
	\item[1] For each $t \in \rho(s)$
    \begin{enumerate}
    	\item[a] Update weights $w_{ts}^{(i)} = w_{ts}^{(i)} \times \phi(X_s=\mu_{ts}^{(i)}, Y_s)$.
        \item[b] Normalize the weights such that $\sum_{i=1}^M w_{ts}^{(i)} = 1$.
    \end{enumerate}
    \item[2] Combine all the incoming messages to form a single set of samples and their weights $\{(w_{s}^{(i)}, \mu_s^{(i)})\}_{i=1}^T$, where $T$ is the sum of all the incoming number of samples.
    \item[3] Normalize the weights such that $\sum_{i=1}^Tw_{s}^{(i)}=1$.
    \item[4] Perform a resampling step to sample new set $\{\mu_s^{(i)}\}_{i=1}^T$ that represent the marginal belief of $X_s$.
\end{enumerate}
\end{AlgoBox}

The pairwise term $\psi_{st}(X_s, X_t)$ can be approximated as the marginal influence function $\zeta(X_t)$ to make the right side of Equation~\ref{eq:7} independent of $X_s$. The marginal influence function provides the influence of $X_s$ for sampling $X_t$. However, this function can be ignored if the pairwise potential function is based on the distance between the variables. This assumption makes Equation~\ref{eq:7} avoid the step of integration and sample $\hat{X}_t^{(i)}$ from the ``pre-message'' followed by a pairwise sampling where $\psi_{st}(X_s, X_t)$ is acting as $\psi_{st}(X_s | X_t=\hat{X}_t^{(i)})$ to get a sample $\hat{X}_s^{(i)}$. To represent message $m_{t\rightarrow s}^n(X_s)$, the $M$ samples $\{\hat{X}_s\}_{i=1}^M$ are considered as $\{\mu_{ts}\}_{i=1}^M$. $\{\Lambda_{ts}\}_{i=1}^M$ are computed using Kernel Density Estimation methods. PAMPAS~\cite{isard2003pampas} has a slightly different notation and methods to compute the samples.



The Gibbs sampling procedure in itself is an iterative procedure and hence makes the computation of the "pre-message" (as the Foundation function described for PAMPAS) expensive as $M$ increases. In the next section, we provide our proposed message representation followed by the algorithm to compute $m_{t\rightarrow s}^n(X_s)$ at iteration $n$.



\section{Nonparametric Belief Propagation using Pull Message Passing}
\label{pmpnbp}

Given the overview of Nonparametric Belief Propagation above in Section~\ref{nbp_overview}, 
we now describe our ``pull'' message passing algorithm. We represent message as a set of pairs instead of triplets in Equation~\ref{eq:3} which is
\begin{equation}
m_{t\rightarrow s} = \{(w_{ts}^{(i)},\mu_{ts}^{(i)}): 1\leq i\leq M\}
\end{equation}
Similarly, the marginal belief is summarized as a sample set
\begin{equation}
bel_s^n(X_s) = \{\mu_{s}^{(i)}: 1\leq i\leq T\}
\end{equation}
where $T$ is the number of samples representing the marginal belief.
We assume that there is a marginal belief over $X_s$ as $bel_s^{n-1}(X_s)$ from the previous iteration. To compute the $m_{t\rightarrow s}^n(X_s)$, at iteration $n$, we initially sample $\{\mu_{ts}^{(i)}\}_{i=1}^M$ from the belief $bel_s^{n-1}(X_s)$. Pass these samples over to the neighboring nodes $\rho(t)\setminus s$ and compute the weights $\{w_{ts}^{(i)}\}_{i=1}^M$. This step is described in \ref{alg:message_update}. The computation of $bel_s^{n}(X_s)$ is described in ~\ref{alg:belief_update}. The key difference between the ``push'' approach of the earlier methods (NBP and PAMPAS)~\cite{sudderth2003nonparametric,isard2003pampas} and our ``pull'' approach is the message $m_{t\rightarrow s}$ generation. In the ``push'' approach, the incoming messages to $t$ determines the outgoing message $t\rightarrow s$. Whereas, in the ``pull'' approach, samples representing $s$ are drawn from its belief $bel_s$ from previous iteration and weighted by the incoming messages to $t$. This weighting strategy is computationally efficient. Additionally, the product of incoming messages to compute $bel_s$ is approximated by a resampling step as described in ~\ref{alg:belief_update}.

\section{Experimental Setup}
\label{experimental_setup}

\begin{figure}[t!]
    \centering
    \captionsetup[subfigure]{labelformat=empty}
    \subfloat[(a) Graphical model]{\includegraphics[width=0.48\columnwidth]{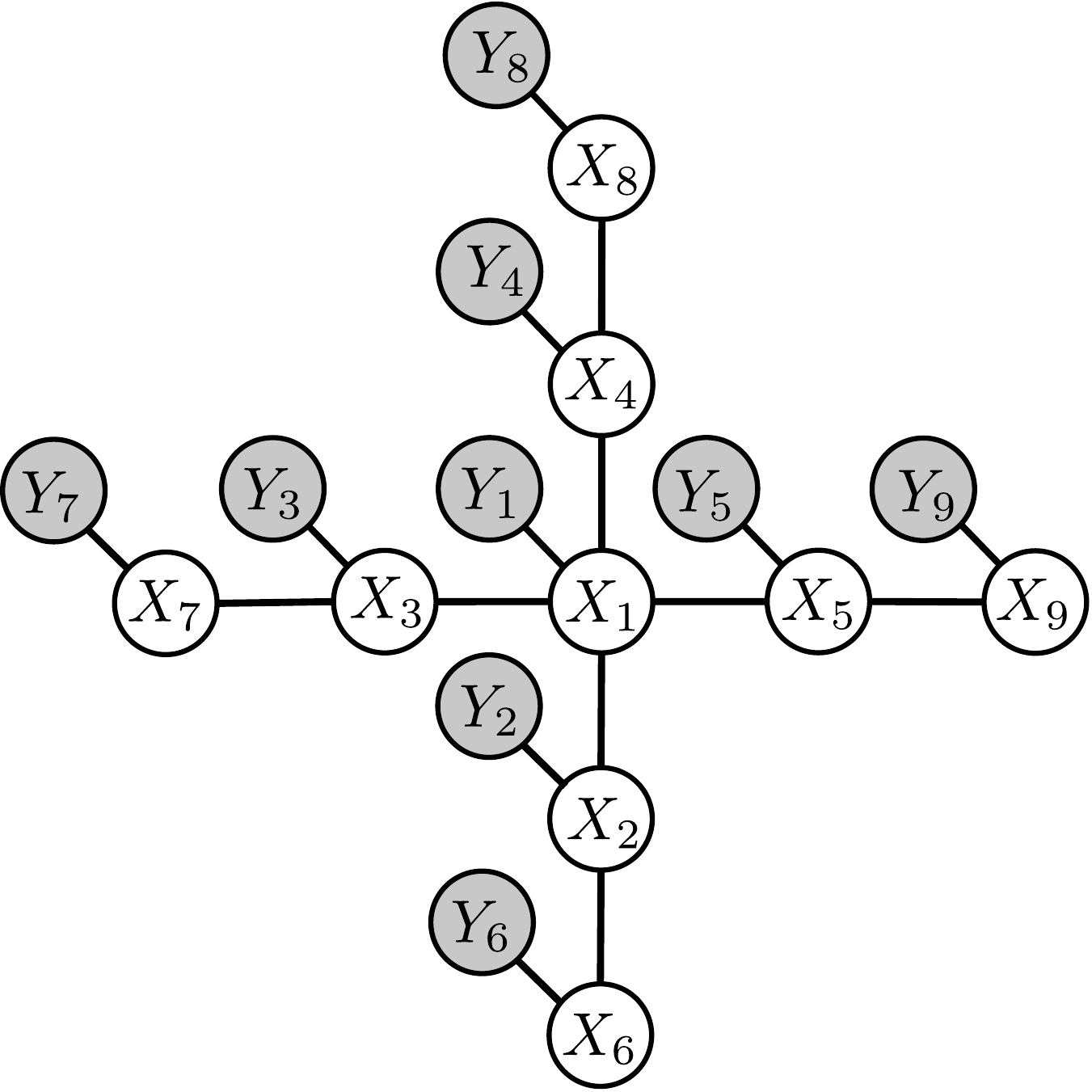}}  ~~ ~~
\subfloat[(b) Geometrical structure]{\includegraphics[width=0.48\columnwidth]{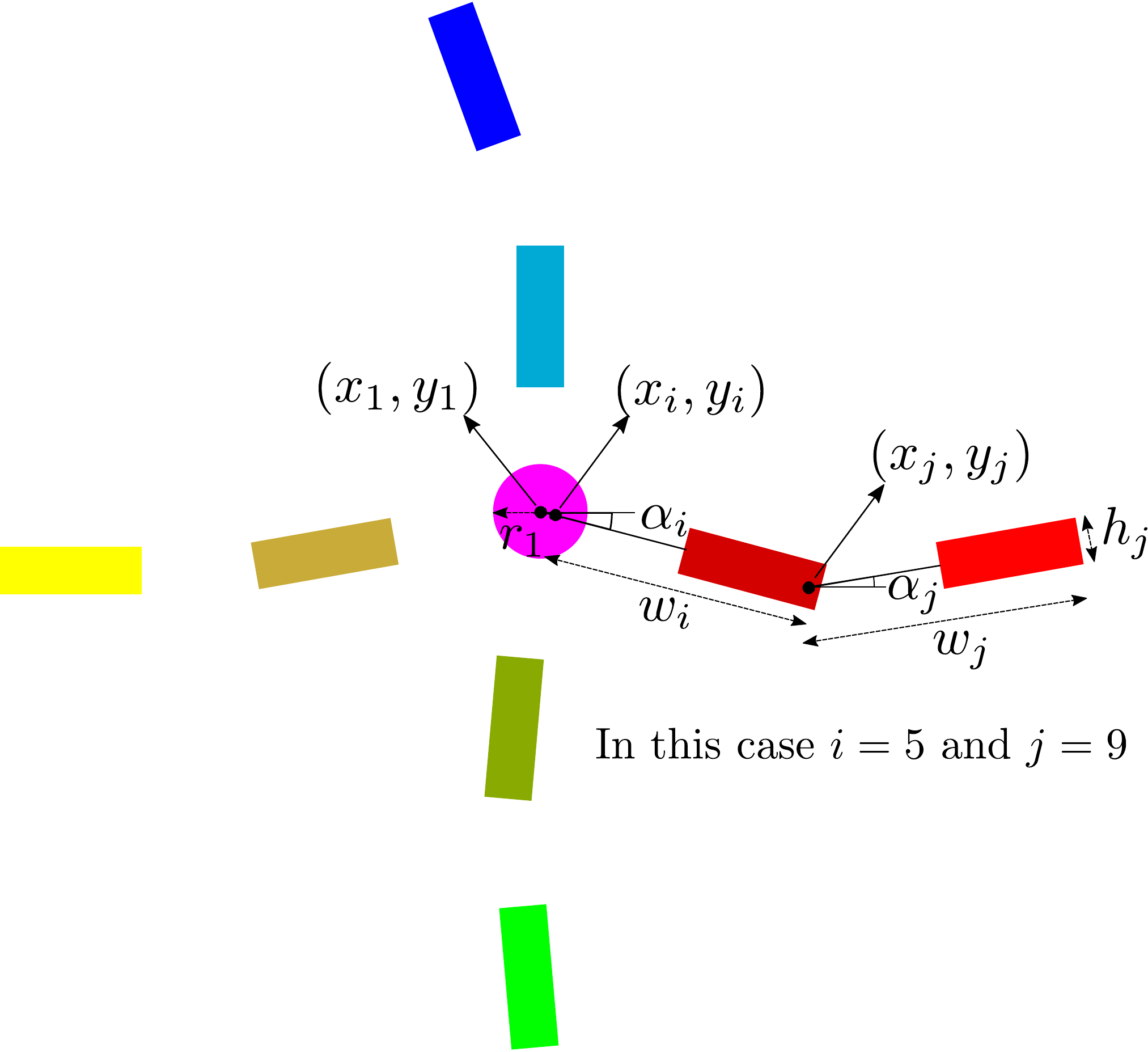}}
 \caption{\footnotesize The pattern used for the experiments has 9 nodes with one circle at the center and four arms with two links each. This forms the graphical model shown in (a), where hidden nodes $X_s$ are connected to their neighbors and informed by observed nodes $Y_s$. Geometrically, the circle and links are defined by their location $(x_s, y_s)$, orientation and dimensions as shown in (b). Color coding here is used to distinguish the links for the qualitative results in the paper.}
 \label{link_structure}
\end{figure}

\begin{figure}[t!]
    \centering
    \captionsetup[subfigure]{labelformat=empty}
    \subfloat[(a) Image observation]{\includegraphics[width=0.20\columnwidth]{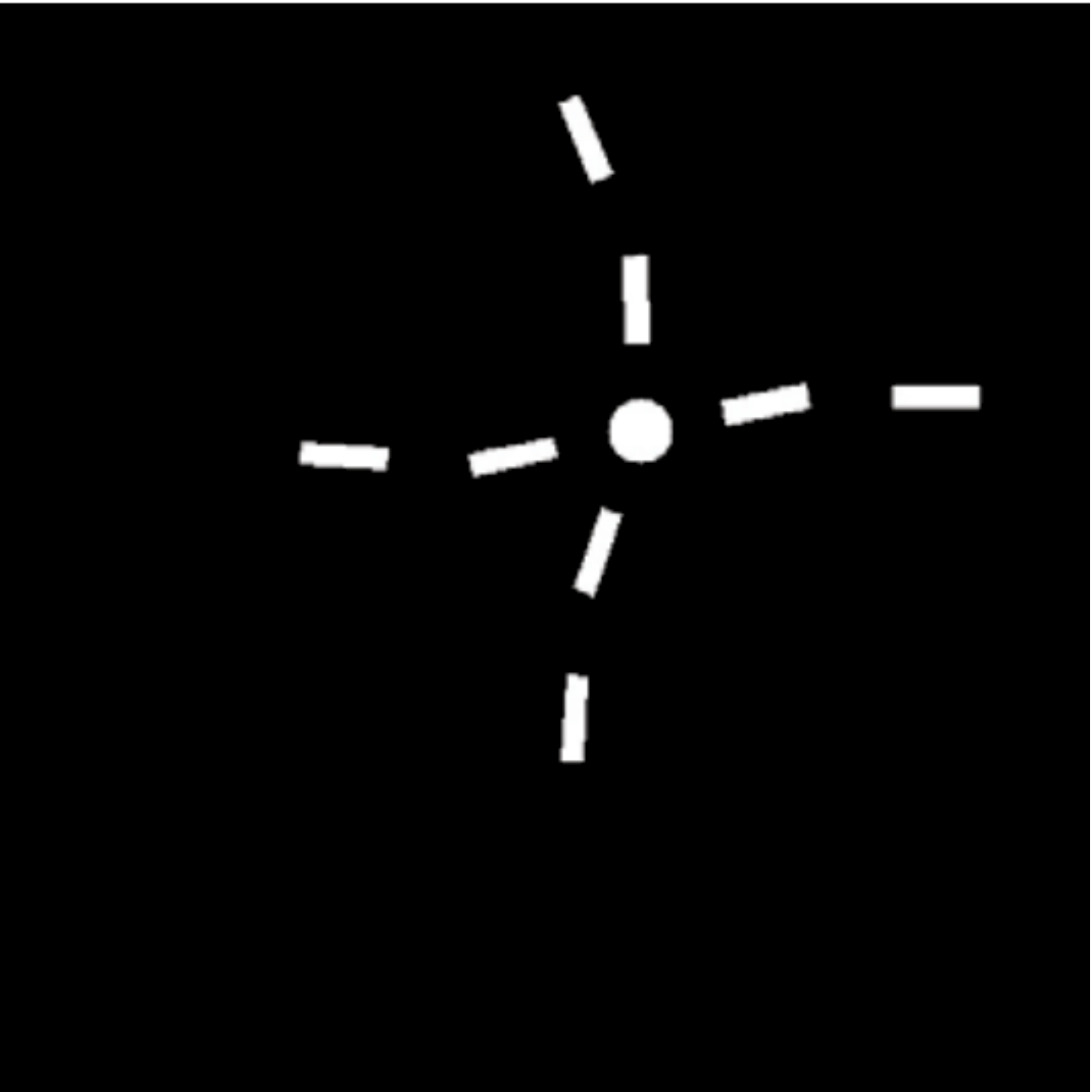}} ~~
\subfloat[(b) $\phi(X_1, Y_1)$]{\includegraphics[width=0.26\columnwidth]{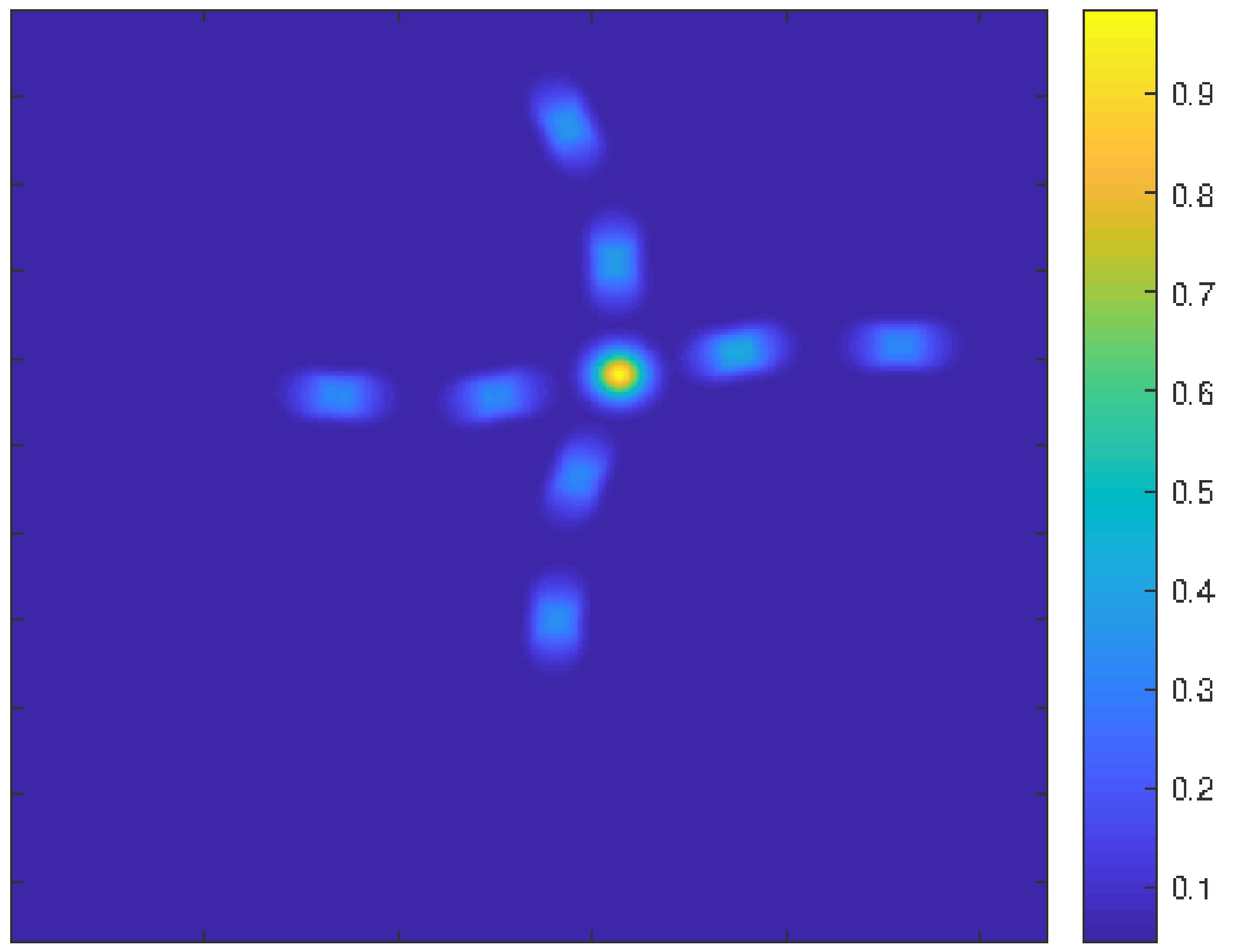}} 
\subfloat[(c) $\phi(X_2, Y_2)$]{\includegraphics[width=0.26\columnwidth]{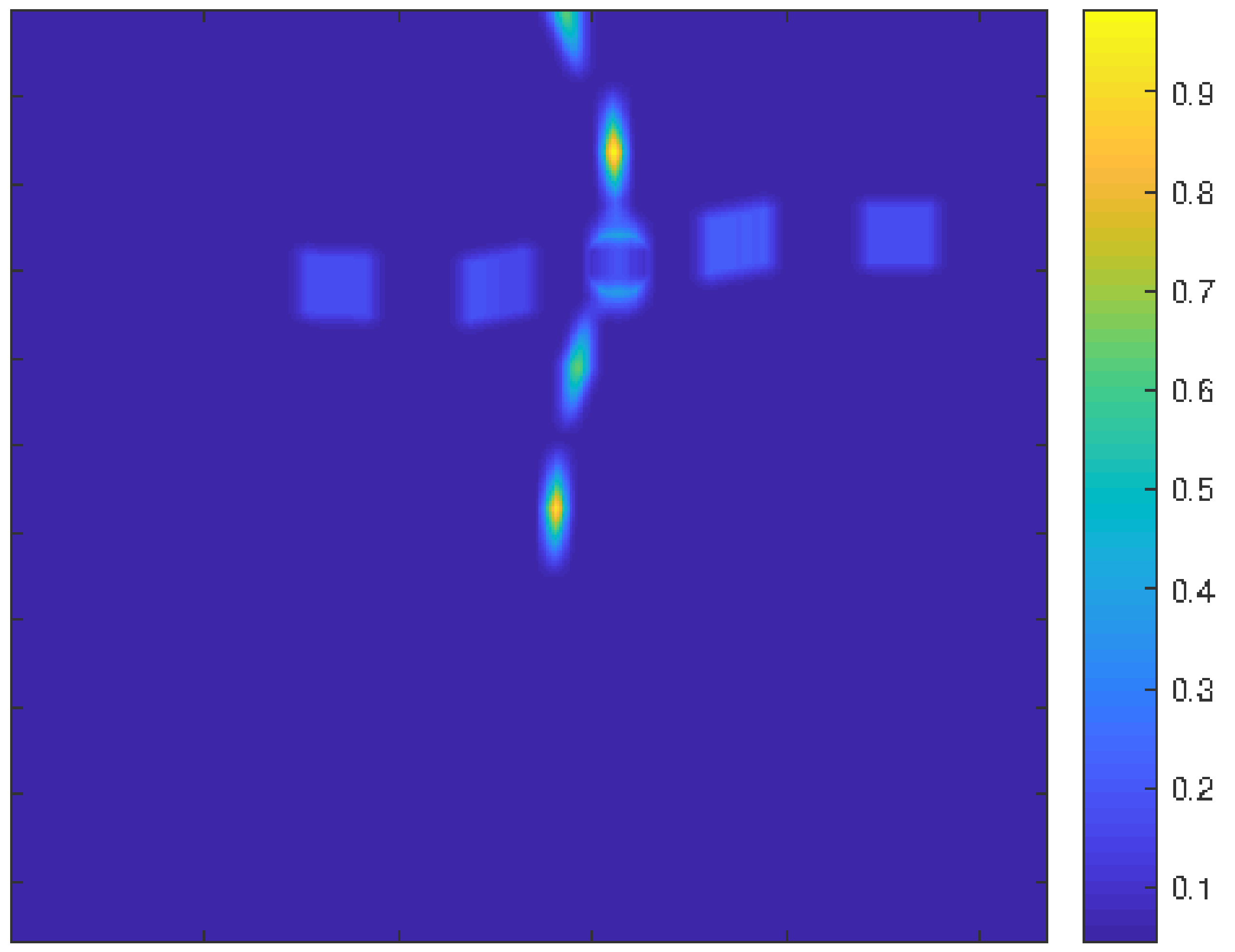}} 
\subfloat[(d) $\phi(X_3, Y_3)$]{\includegraphics[width=0.26\columnwidth]{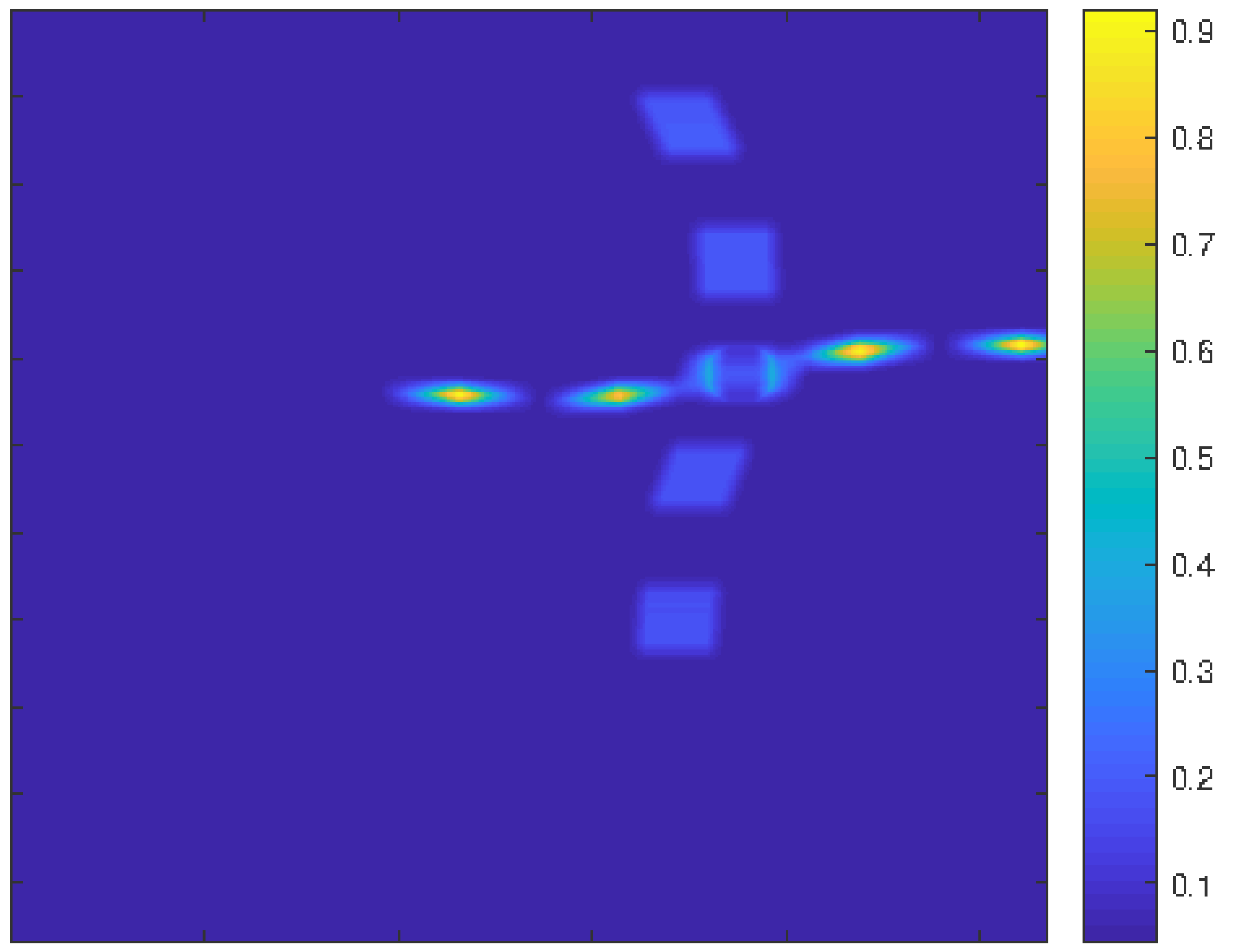}}
 \caption{\footnotesize a) Shows the actual pattern used in the experiments of the paper. b-c) shows the unary potential $\phi(X_s,Y_s)$ for $s=\{1, 2, 3\}$ (circle, vertical rectangular link and horizontal rectangular link respectively) with $(x_s, y_s)$ taking all the pixels in the image (a). For ease of understanding, the orientation of the nodes in this illustration are set to $\alpha_1=0$, $\alpha_2=\pi/2$ and $\alpha_3=\pi$.}
 \label{unarypotential}
\end{figure}

We compare our proposed PMPNBP method with PAMPAS~\cite{isard2003pampas} on their 2D illustratory example (Figure~\ref{link_structure}). The pattern has circle node with state variable $X_1=(x_1, y_1, r_1)$ denoting its position in the 2D image and the radius of the circle. This circle node has four arms with two links each. These links are nodes in the graph with state variables $X_i(x_i, y_i, \alpha_i, w_i, h_i)$. The links connected to the circle are indexed as $2\leq i \leq 5$ with their connected outer links as $j=i+4$. In the recreation of this illustratory example, we define the unary potential as
\begin{equation}
\phi(X_s, Y_s) = \begin{cases}
1-\frac{\abs{I_{sub}(\{(x_s, y_s)\}_{p=1}^P)-T(\{(x_s, y_s)\}_{p=1}^P)}}{\max(P, Q)} &X_s \in \text{circle}\\
1-\frac{\abs{I_{sub}(\{(x_s, y_s)\}_{p=1}^P)-T(\{(x_s, y_s, \alpha_s, w_s, h_s)\}_{p=1}^P)}}{\max(P, Q)} &X_s \in \text{links}
\end{cases}
\end{equation}

where $I_{sub}$ is the patch of image centered at $(x_s, y_s)$ with the same size as the template image $T$ rendered with state of the nodes (circle/links). $P$ and $Q$ are the number of white/observed pixel locations $\{(x, y)\}$ in $I_{sub}$ and $T$ respectively. Figure~\ref{unarypotential} illustrates the computation of the unary potential for nodes $X_1$, $X_2$, $X_3$ visually.

\begin{figure}[t!]
    \centering
    \captionsetup[subfigure]{labelformat=empty}
\subfloat[(a) $\psi(X_9|X_5=\mu_5)$]{\includegraphics[width=0.24\columnwidth]{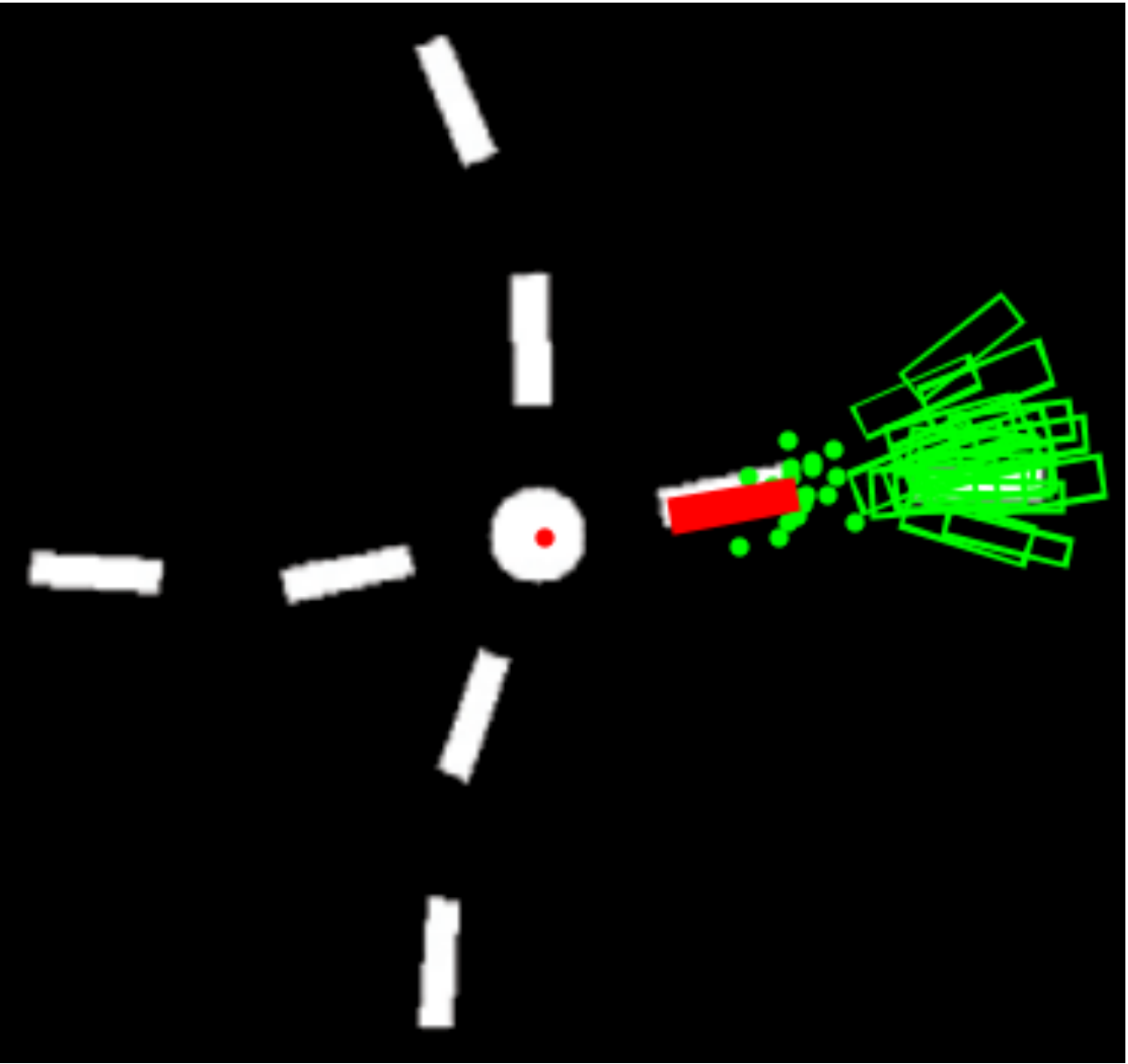}} ~~
\subfloat[(b) $\psi(X_5|X_9=\mu_9)$]{\includegraphics[width=0.24\columnwidth]{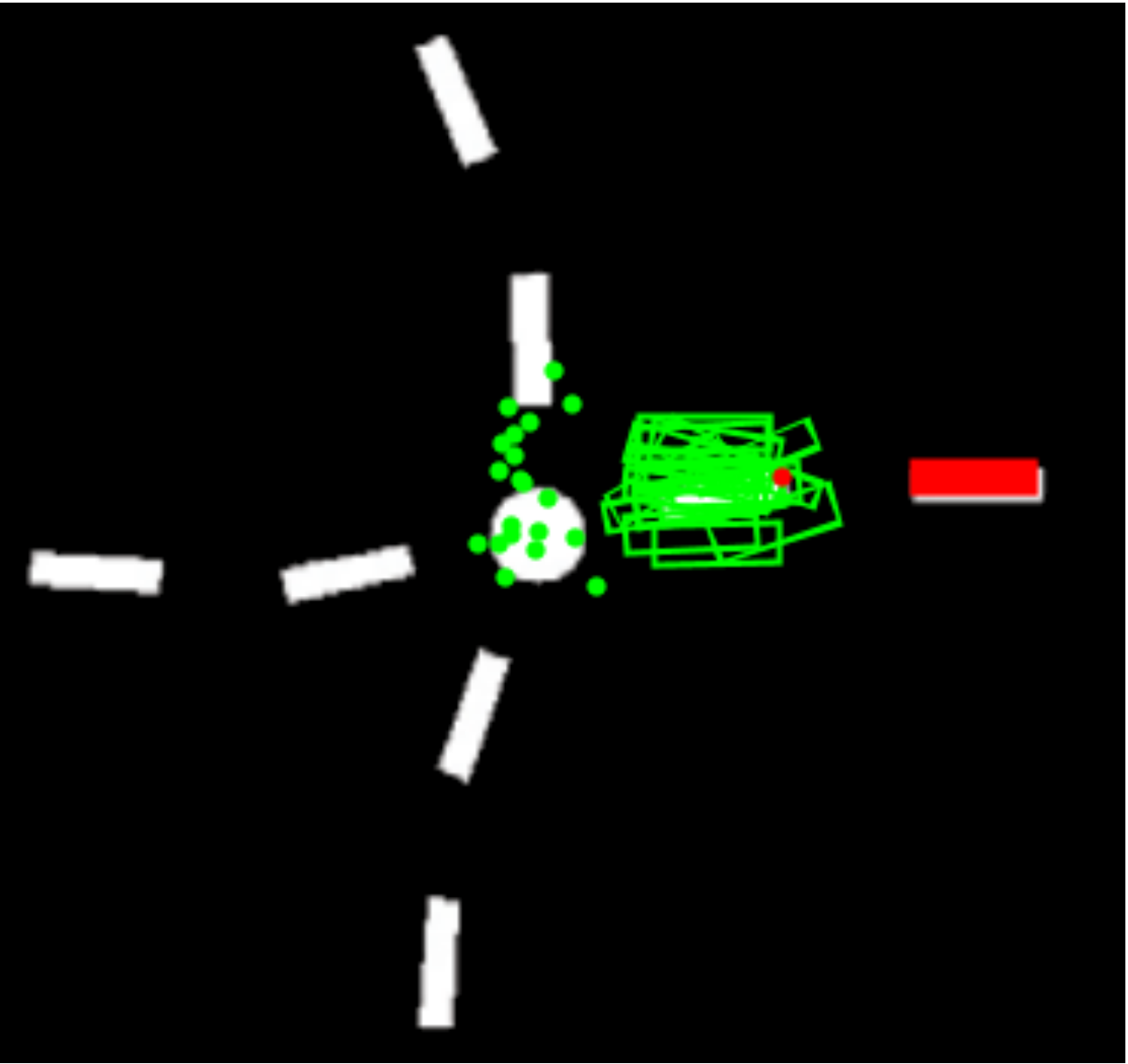}} ~~
\subfloat[(c) $\psi(X_5|X_1=\mu_1)$]{\includegraphics[width=0.24\columnwidth]{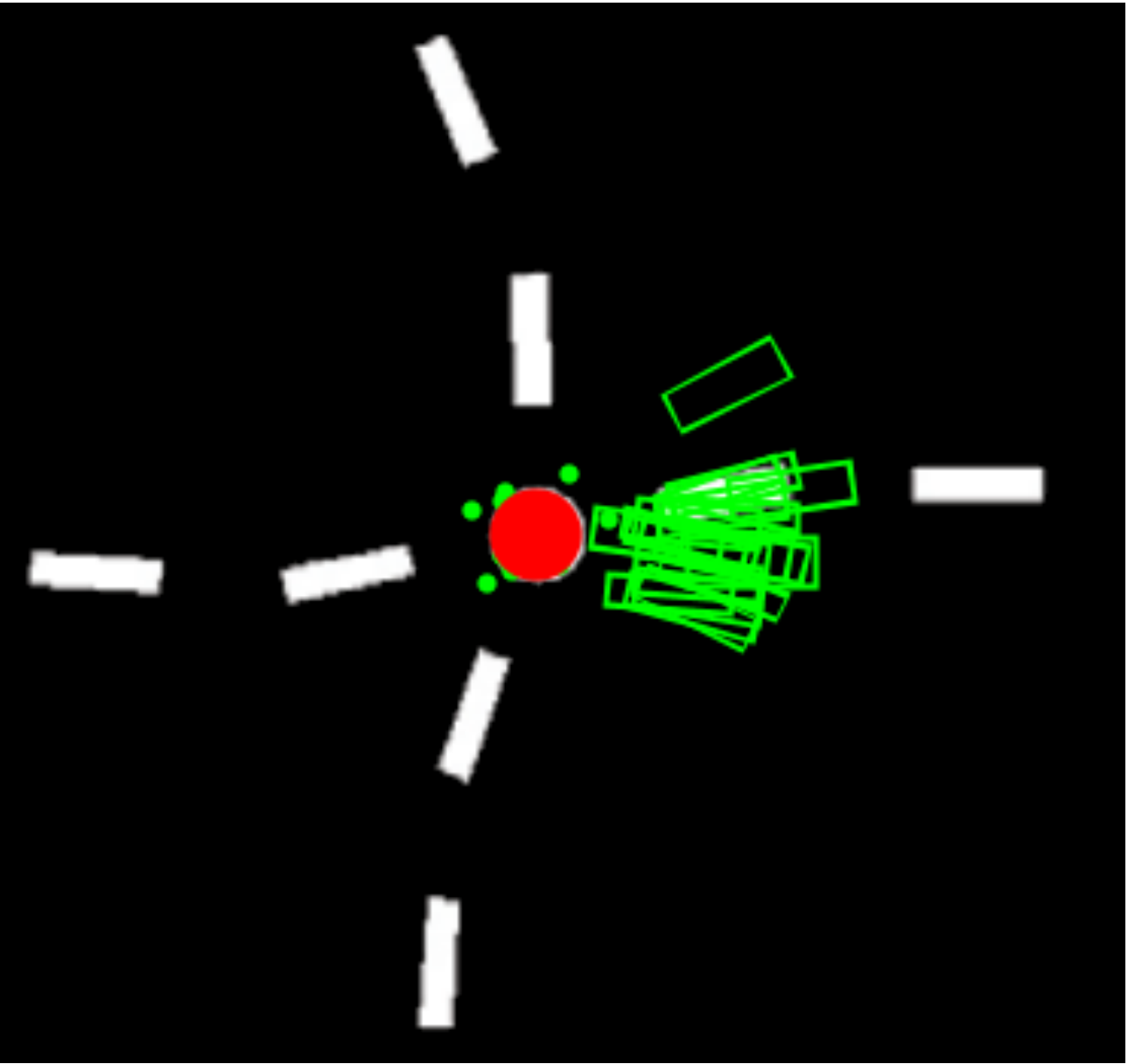}} ~~
\subfloat[(d) $\psi(X_1|X_5=\mu_5)$]{\includegraphics[width=0.24\columnwidth]{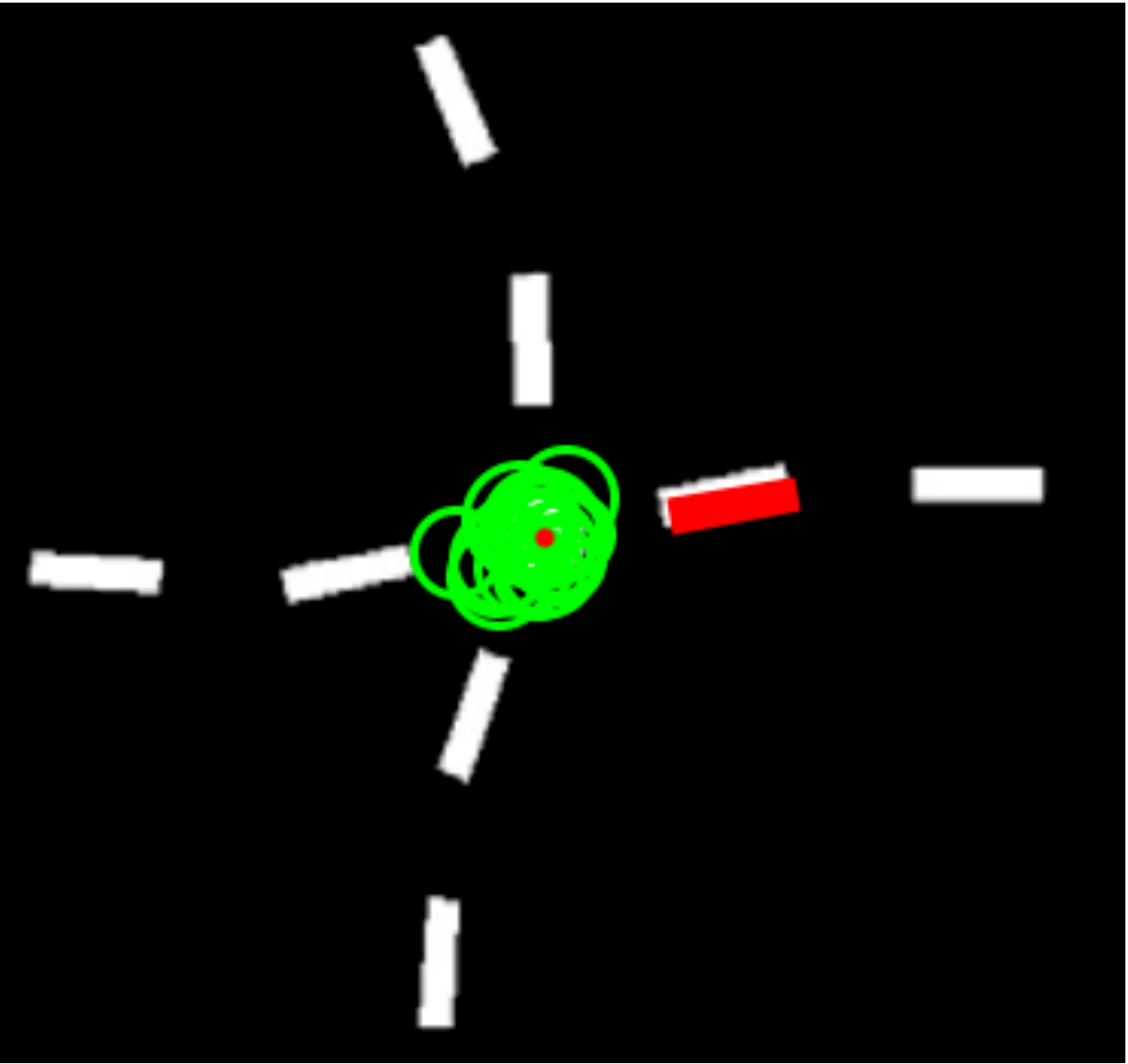}}
 \caption{\footnotesize This figure shows sampling the neighbors based on a given current node sample. For illustration we show the relation between nodes $X_1$, $X_5$ and $X_9$. Each sub-figure shows 20 samples (green color) drawn given its neighboring node (red color) at its ground truth location, constrained by their geometrical relationship.}
 \label{pairwise_sampling}
\end{figure}
The pairwise sampling is done similar to the original description in PAMPAS ~\cite{isard2003pampas}. The procedure to generate samples is described in Appendix A. Figure~\ref{pairwise_sampling} visually illustrates the pairwise sampling for nodes $X_1$, $X_5$, $X_9$.  
With the unary potential and pairwise sampling, we perform inference and report their convergence over iterations in the next section. 

Our implementation of PAMPAS and PMPNBP is in Matlab on a Ubuntu 14.04 Linux machine. A CPU with Core i7 6700HQ - 16 GB RAM is used for all the experiments. Implementation does not involve any type of parallelization to avoid bias in comparisons.

\section{Results} 
\label{results}
\begin{figure}[t!]
    \centering
    \captionsetup[subfigure]{labelformat=empty}
    \subfloat[Belief at iteration 0]{\includegraphics[width=0.24\columnwidth]{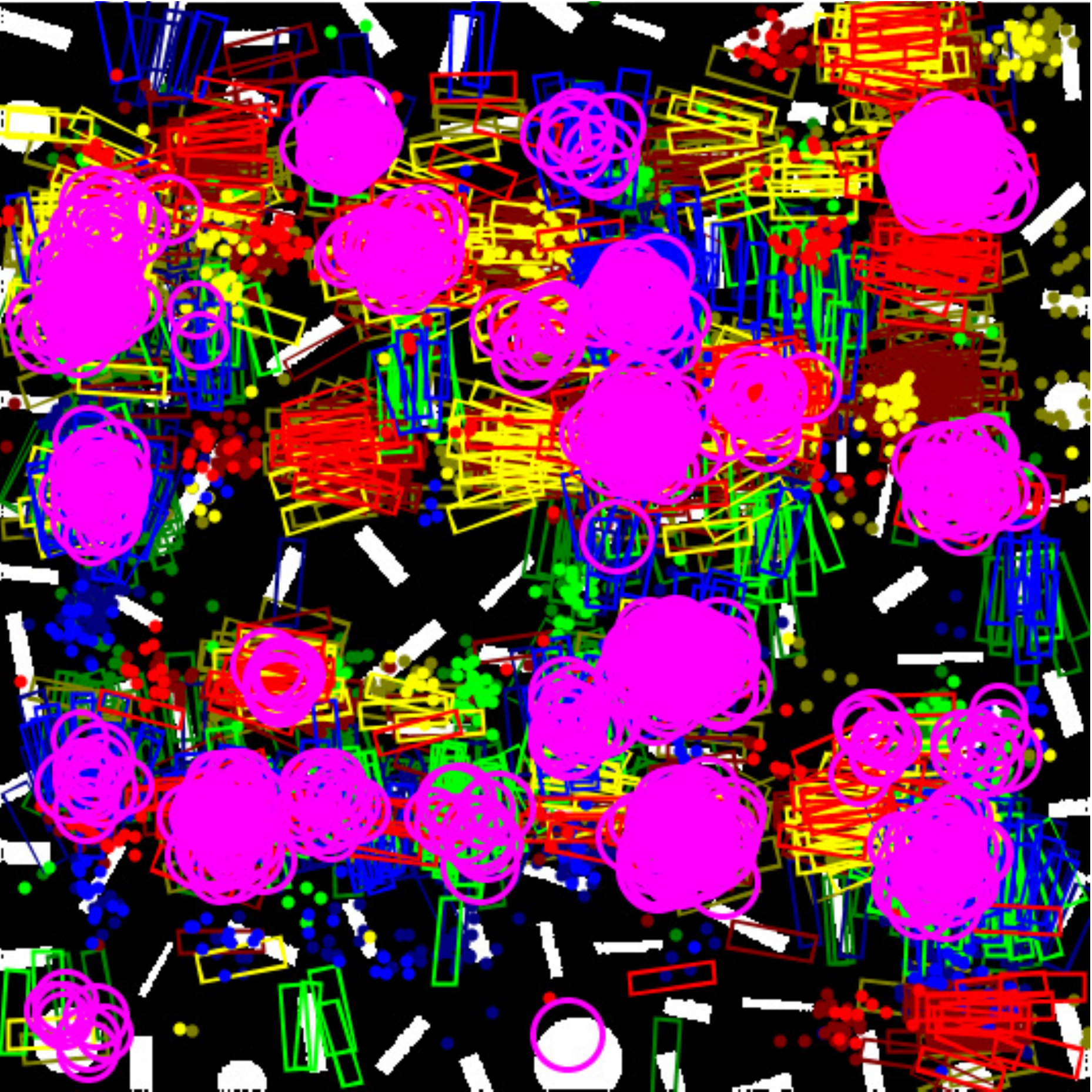}} ~~
\subfloat[Belief at iteration 1]{\includegraphics[width=0.24\columnwidth]{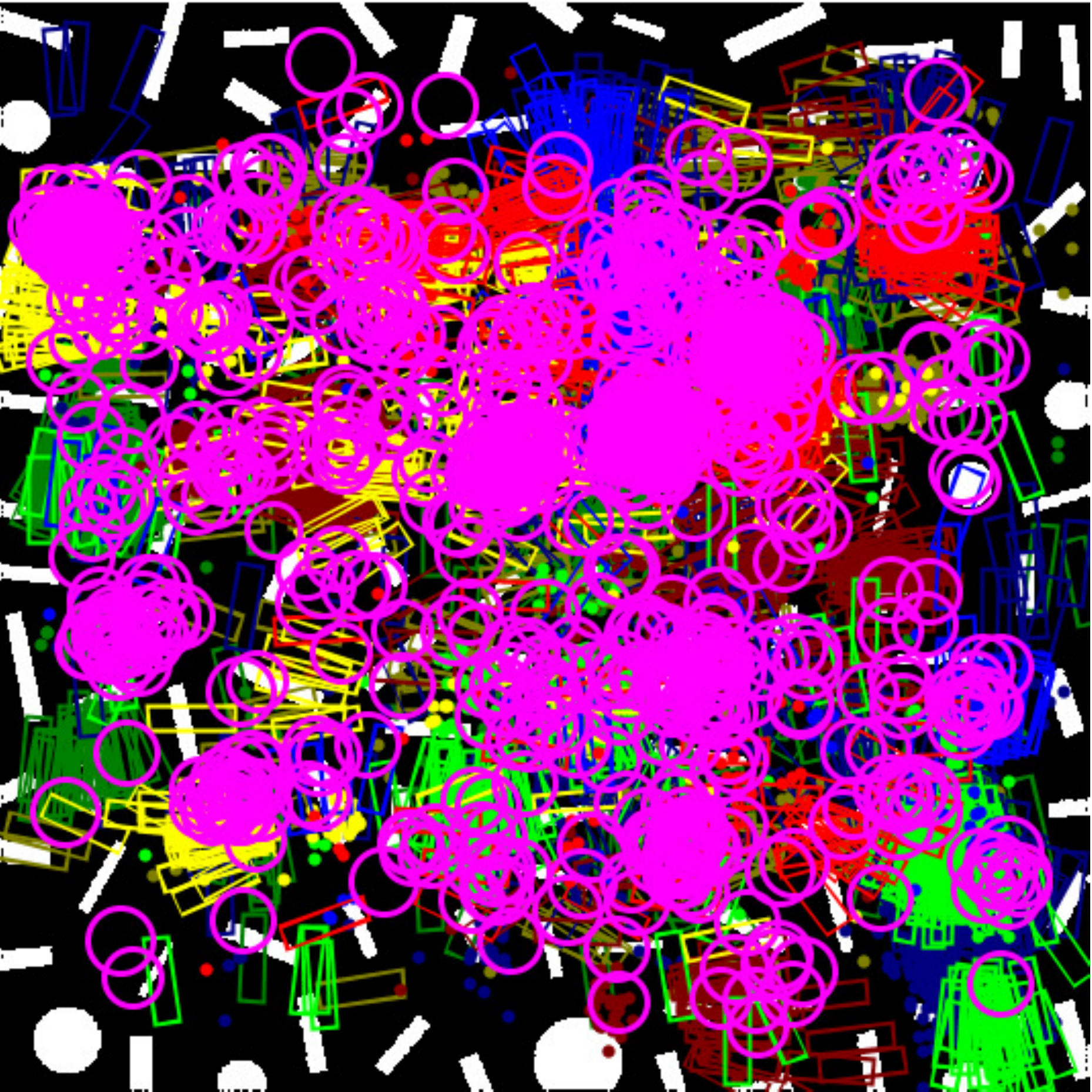}} ~~
\subfloat[Belief at iteration 10]{\includegraphics[width=0.24\columnwidth]{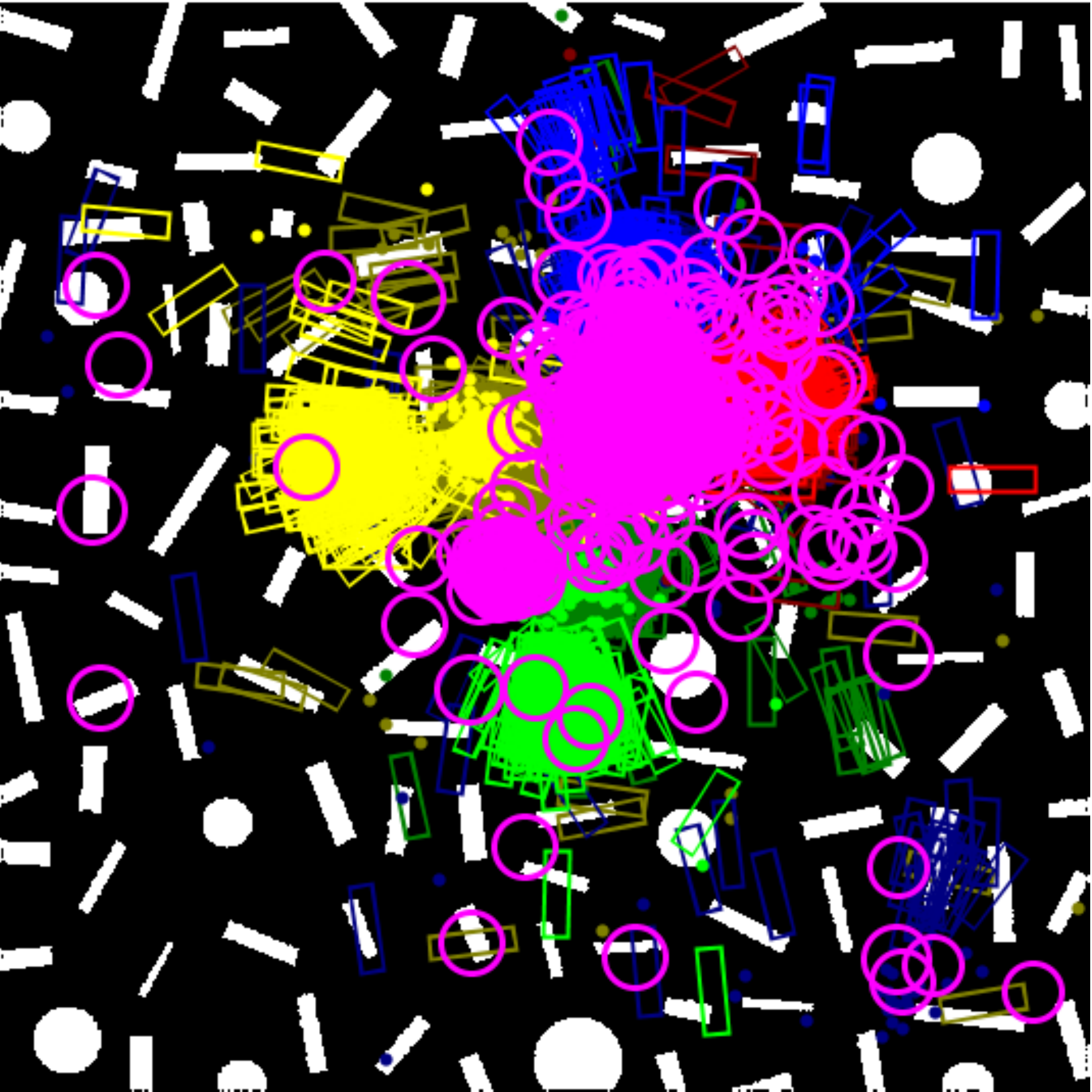}} ~~
\subfloat[Belief at iteration 24]{\includegraphics[width=0.24\columnwidth]{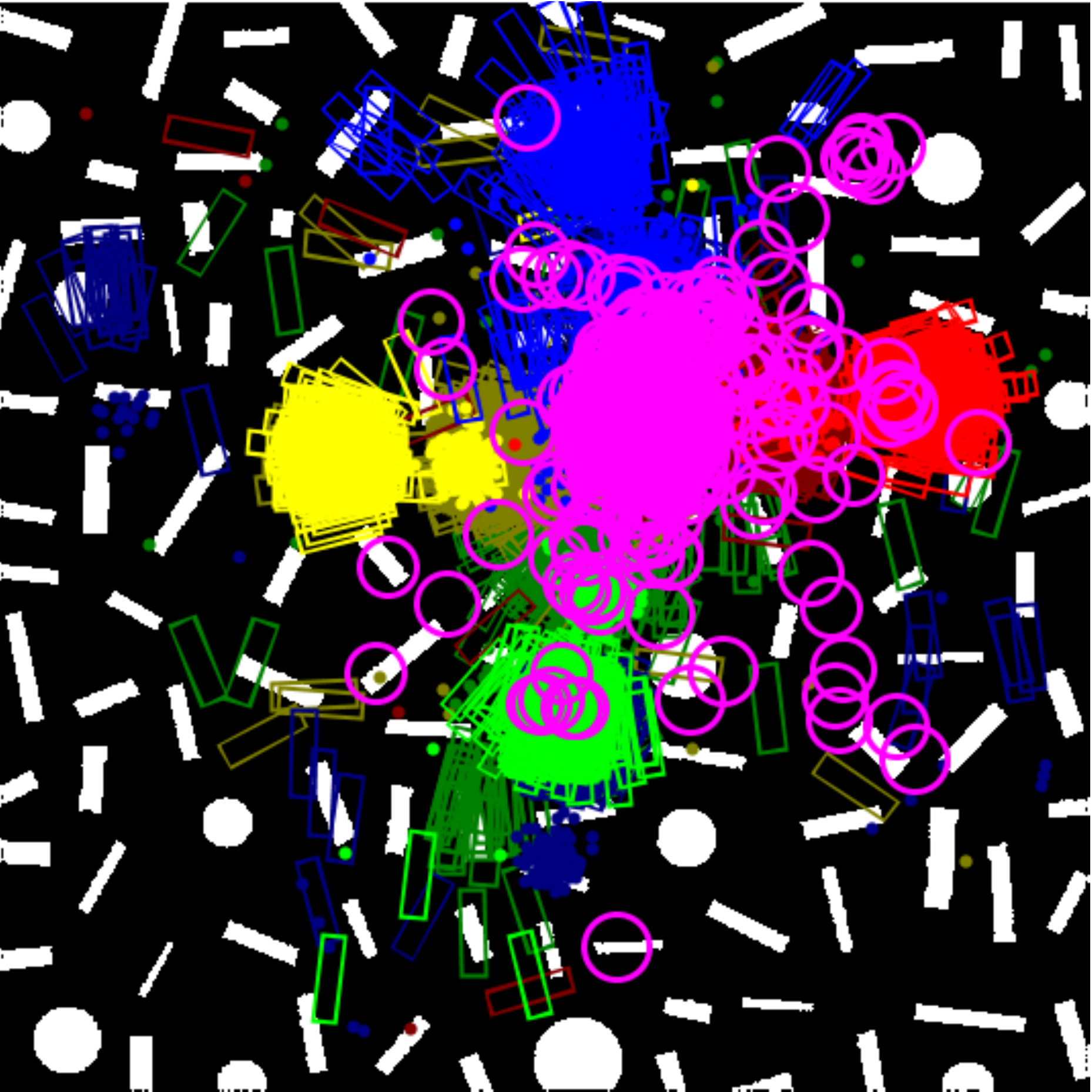}} \\
\subfloat[MLE at iteration 0]{\includegraphics[width=0.24\columnwidth]{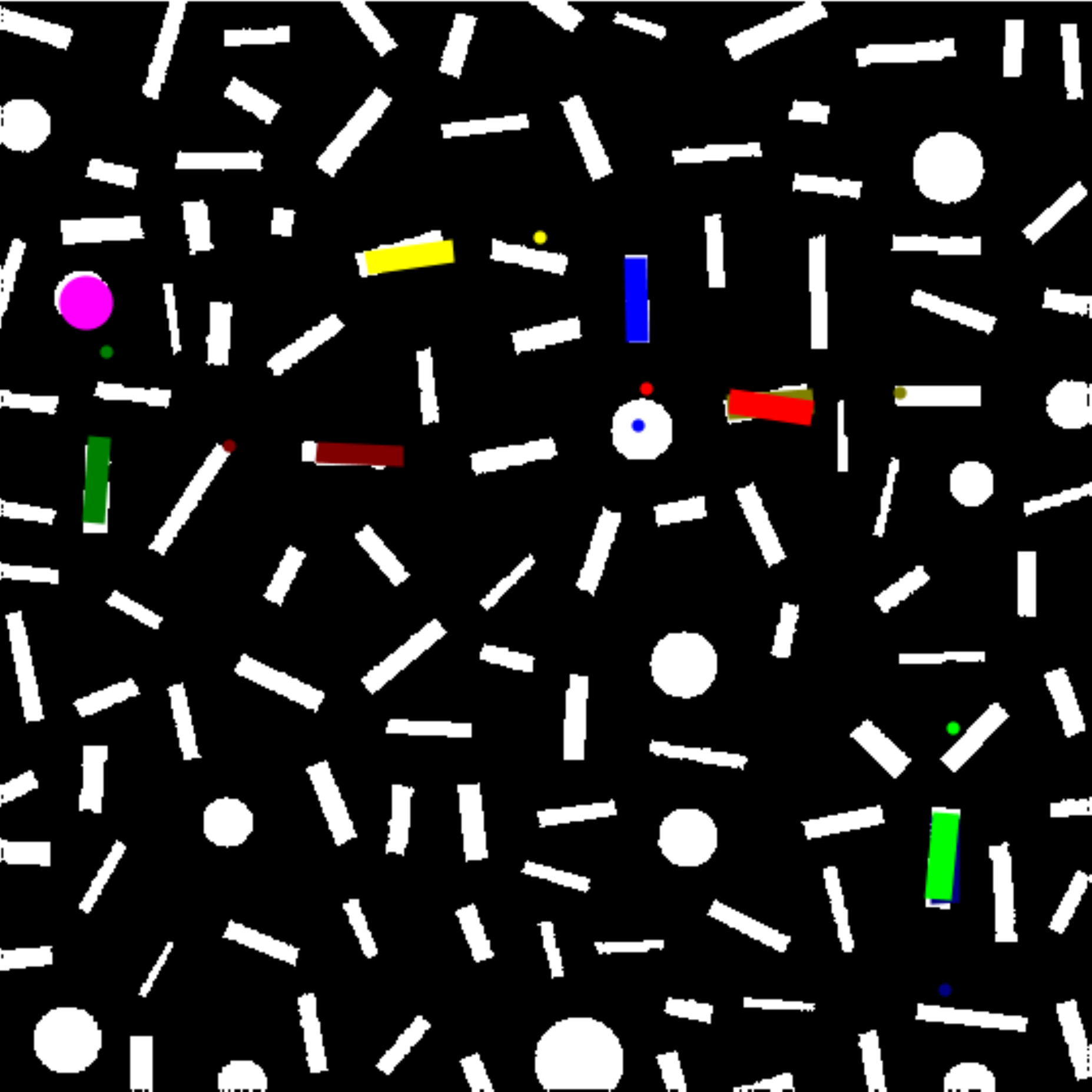}} ~~
\subfloat[MLE at iteration 1]{\includegraphics[width=0.24\columnwidth]{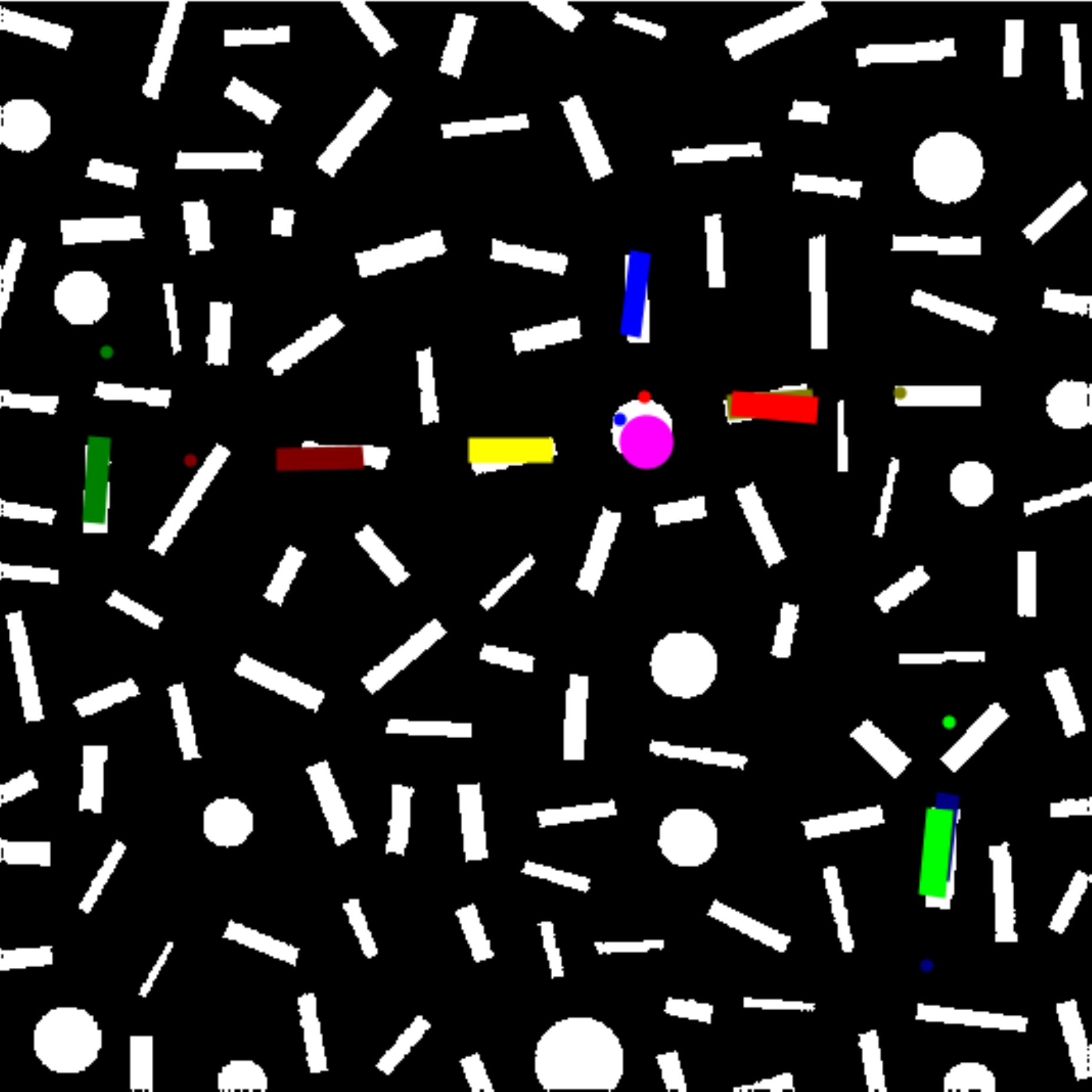}} ~~
\subfloat[MLE at iteration 10]{\includegraphics[width=0.24\columnwidth]{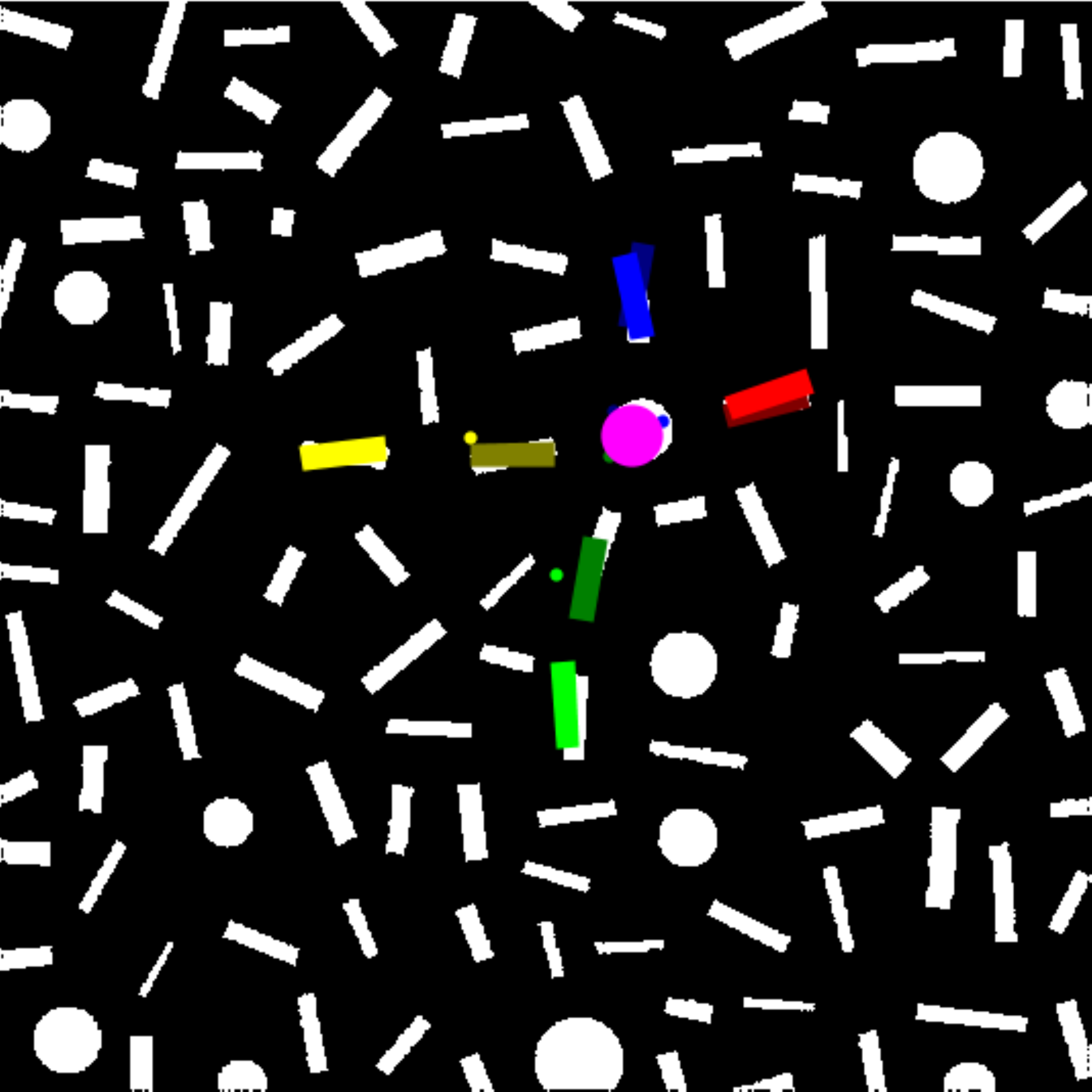}} ~~
\subfloat[MLE at Iteration 24]{\includegraphics[width=0.24\columnwidth]{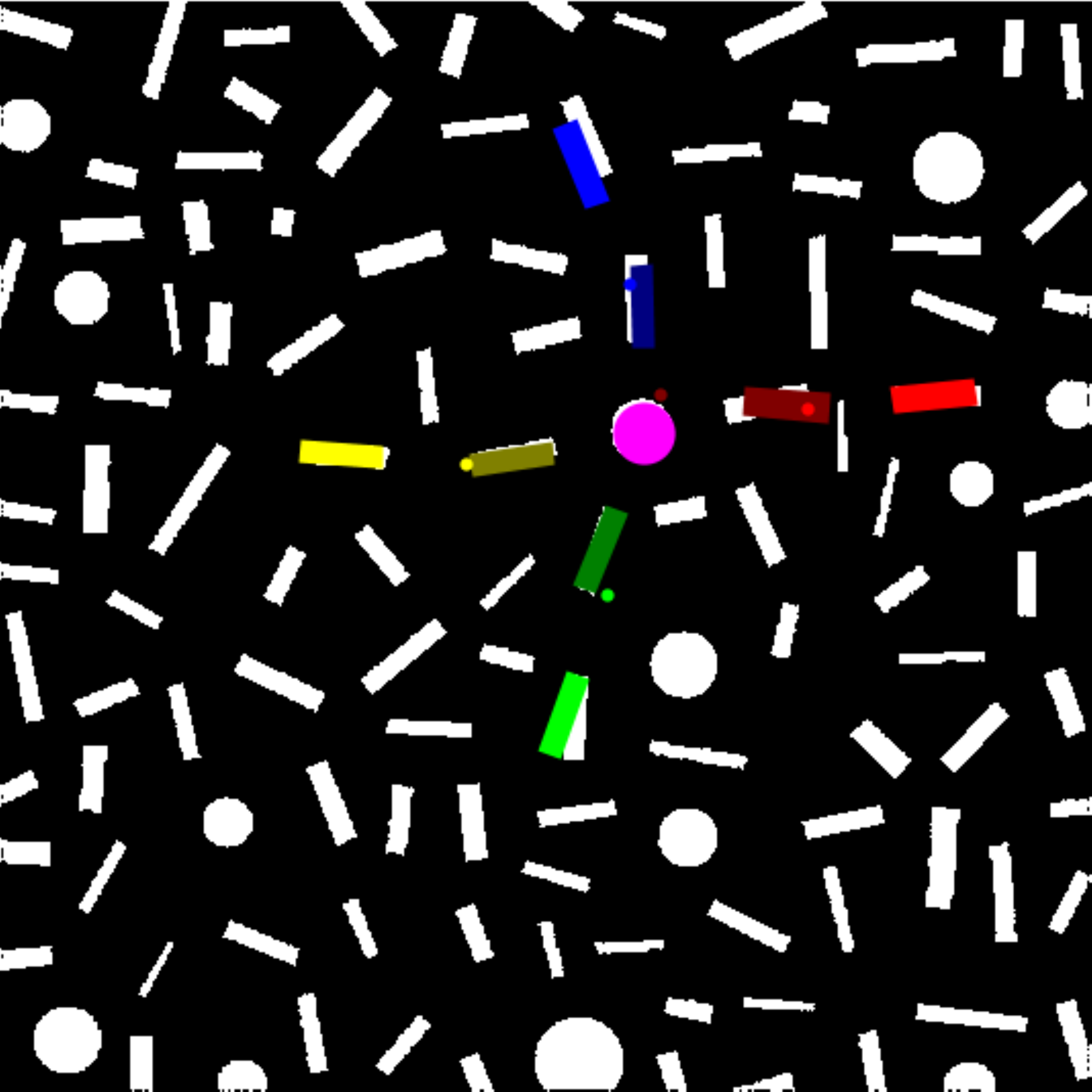}} \\
 \caption{\footnotesize PMPNBP results with circle node observed. Each message contains 200 particles initialized randomly at locations where their $\phi_s > 0.4$. The top row shows the belief samples $bel_s$ for each of the nodes and the bottom row shows their Maximum Likely Estimate (MLE). MLE at iteration 24 has all the links and circle converged to their ground truth states (Best viewed in color).}
 \label{pmpnbp_results_circle}
\end{figure}

\begin{figure}[t!]
    \centering
    \captionsetup[subfigure]{labelformat=empty}
    \subfloat[Belief at iteration 0]{\includegraphics[width=0.24\columnwidth]{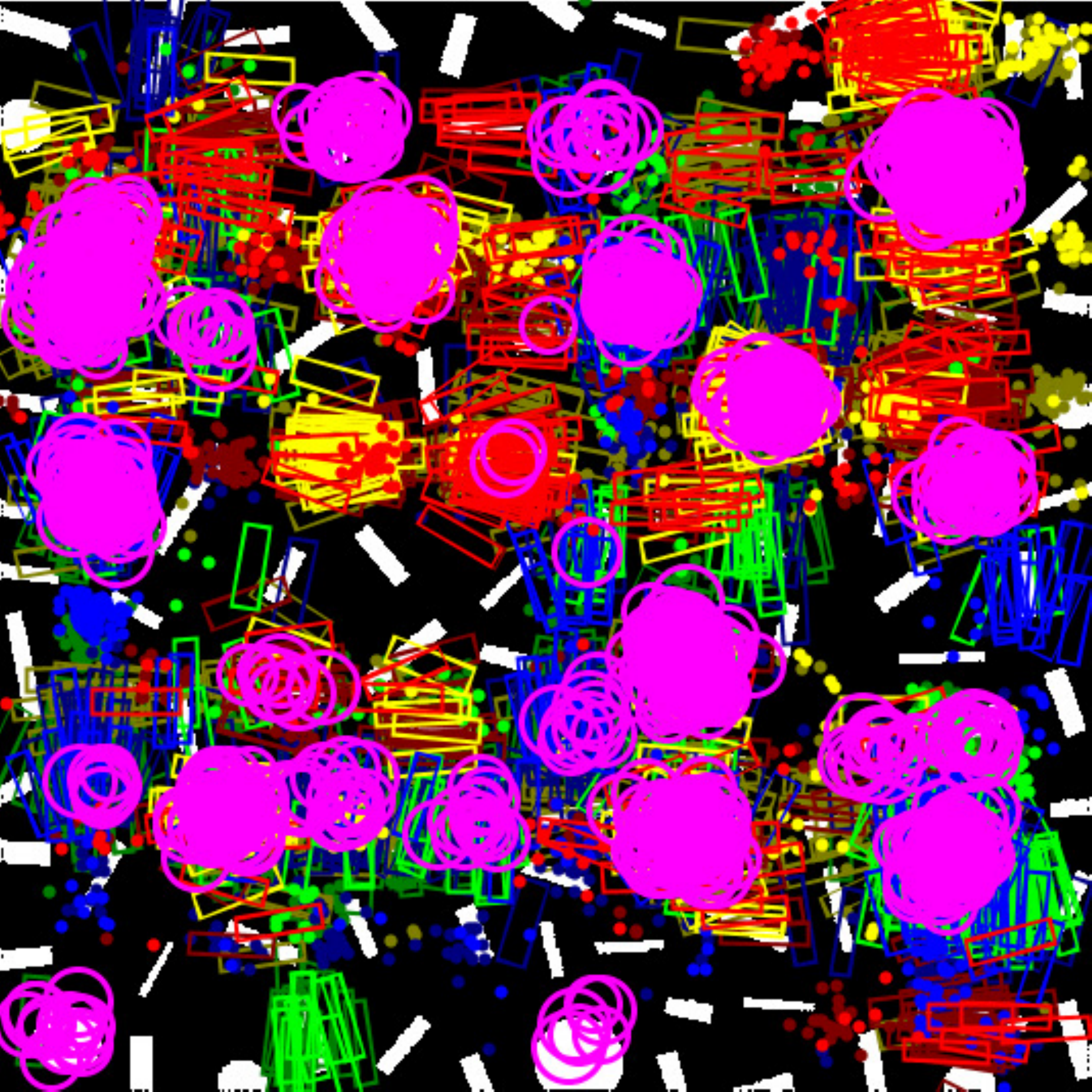}} ~~
\subfloat[Belief at iteration 1]{\includegraphics[width=0.24\columnwidth]{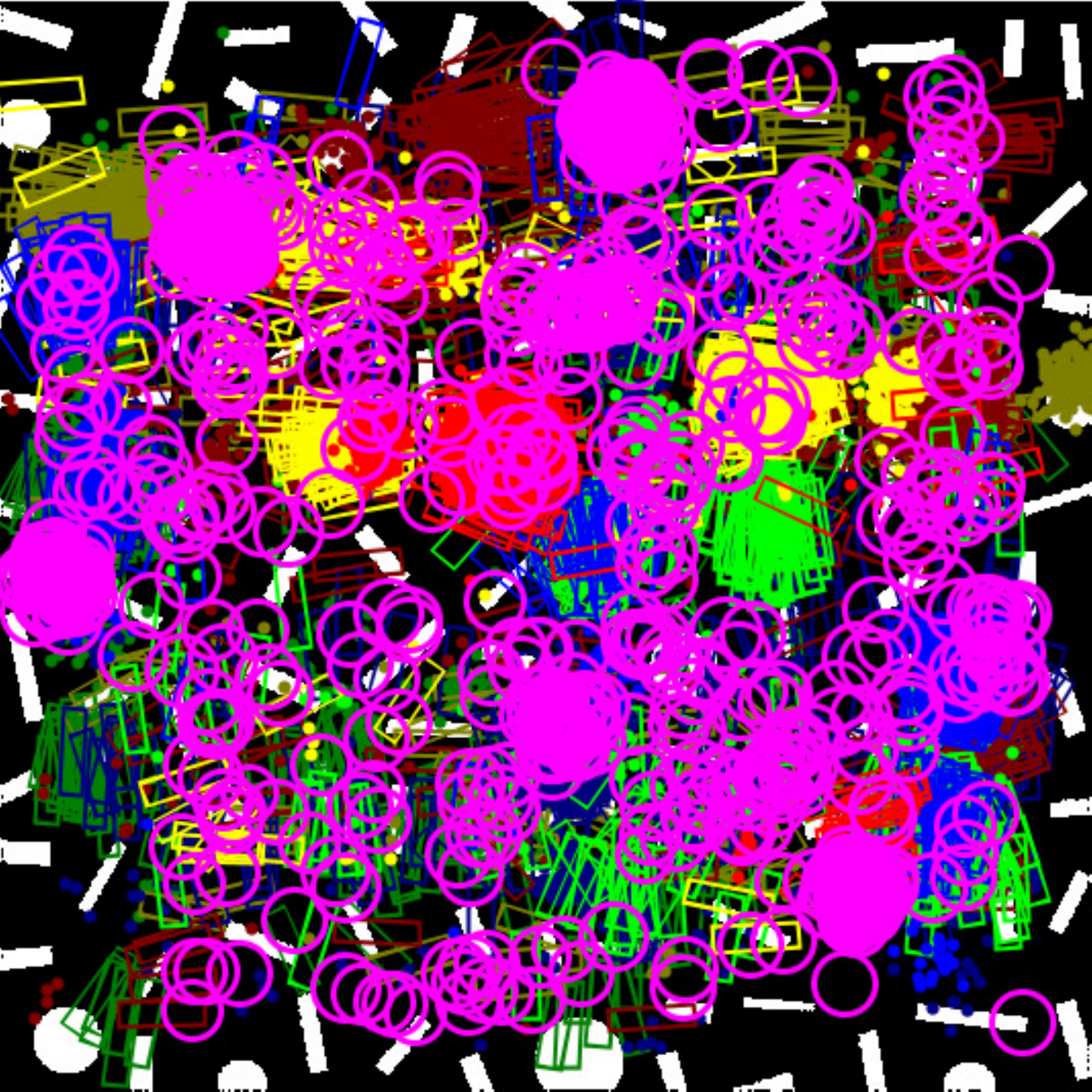}} ~~
\subfloat[Belief at iteration 10]{\includegraphics[width=0.24\columnwidth]{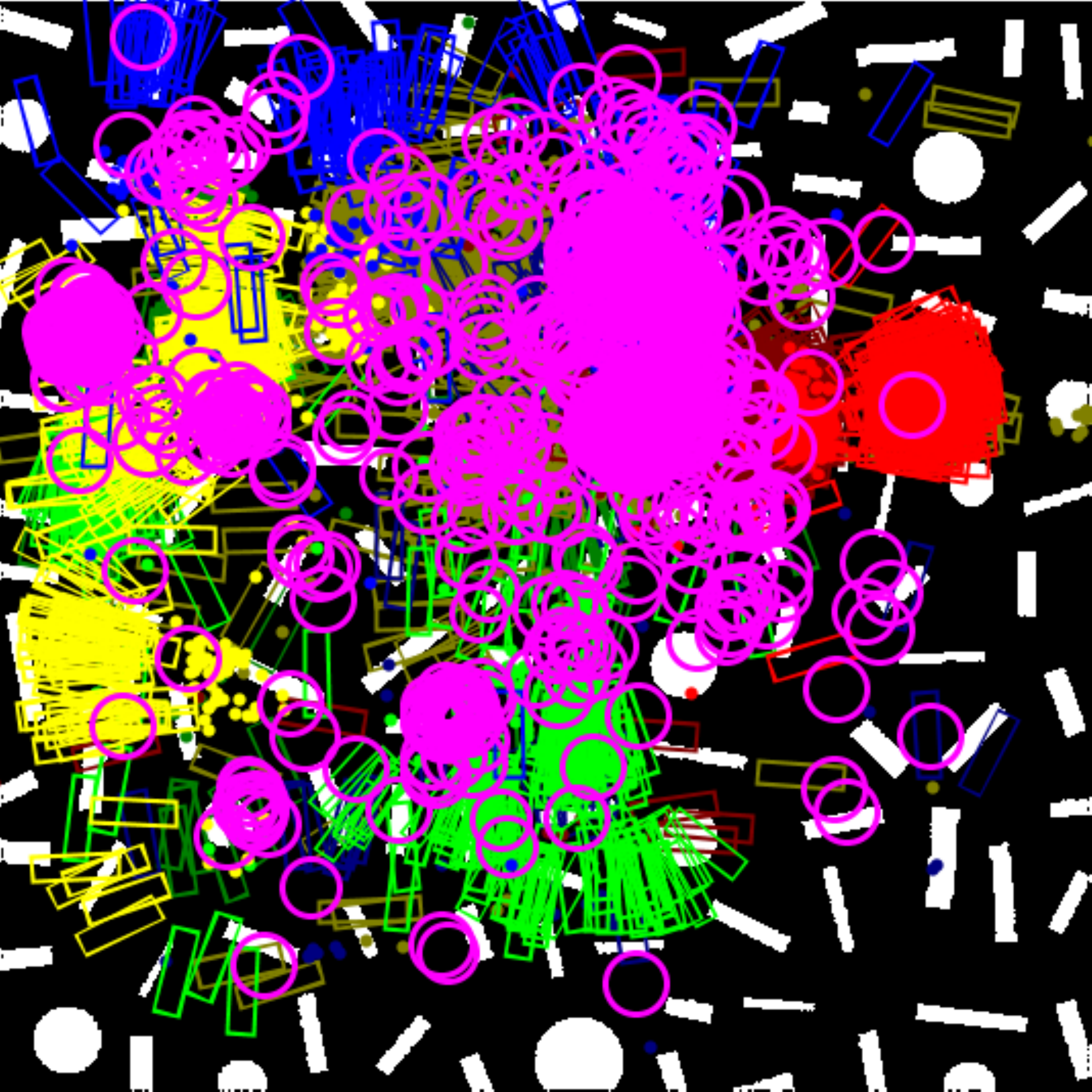}} ~~
\subfloat[Belief at iteration 34]{\includegraphics[width=0.24\columnwidth]{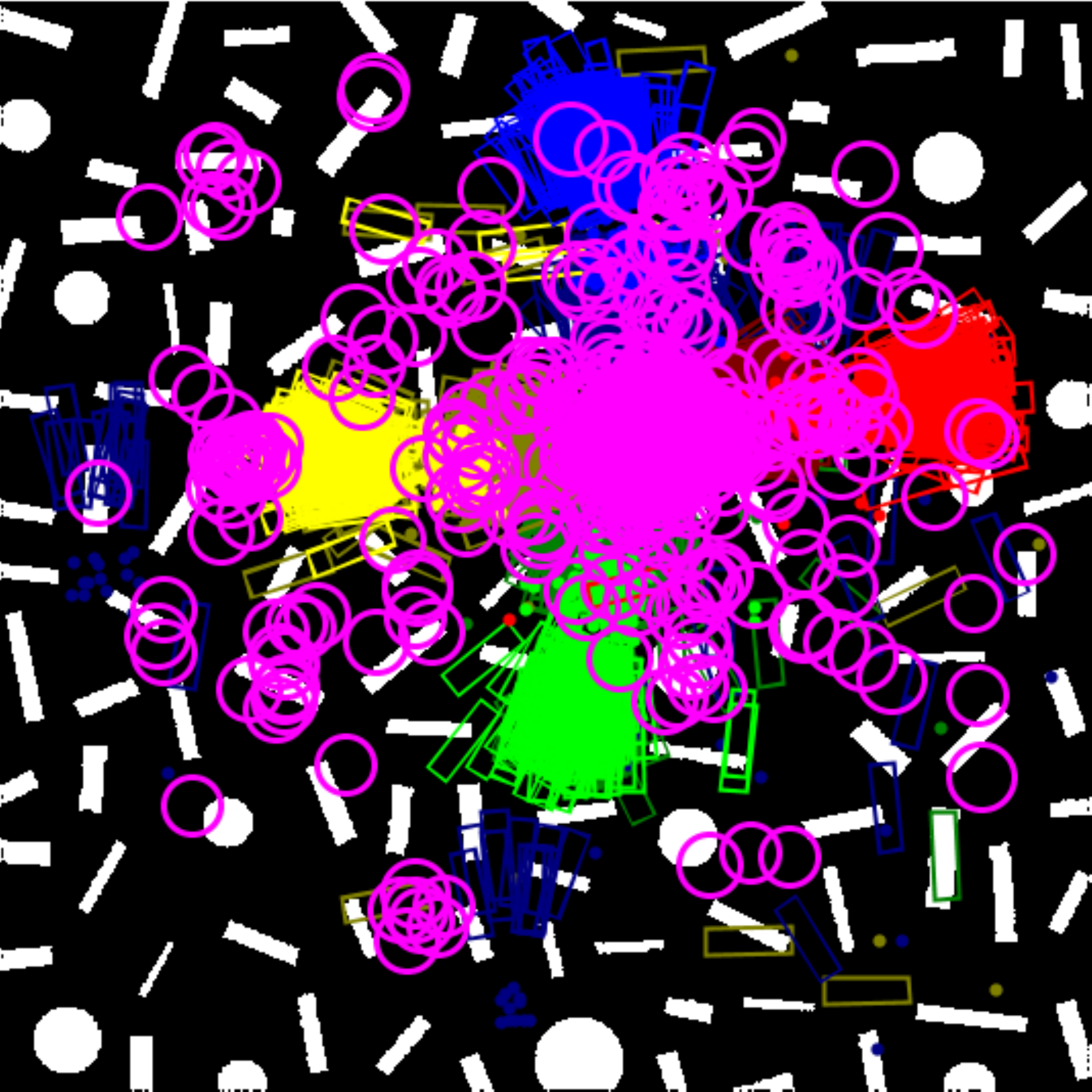}} \\
\subfloat[MLE at iteration 0]{\includegraphics[width=0.24\columnwidth]{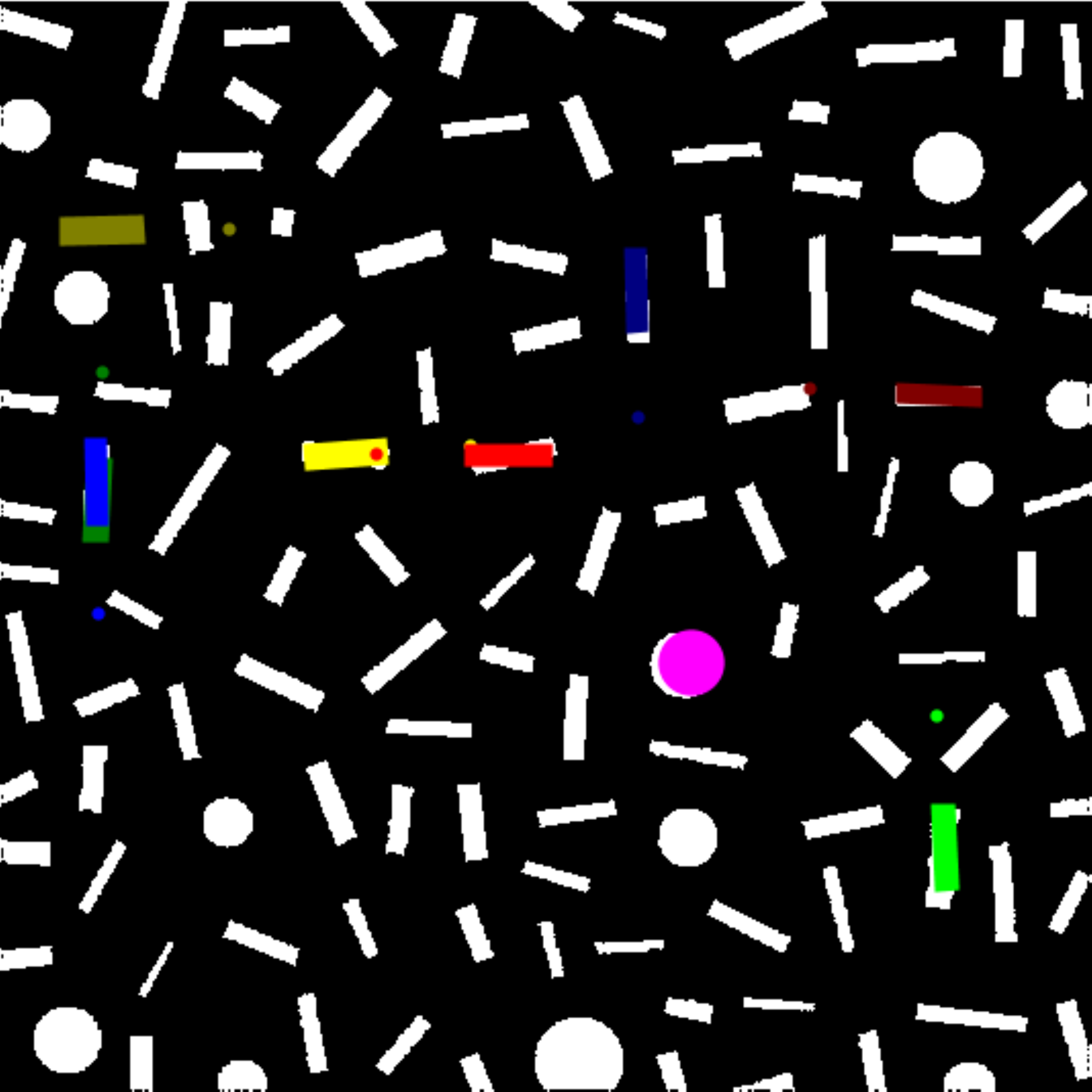}} ~~
\subfloat[MLE at iteration 1]{\includegraphics[width=0.24\columnwidth]{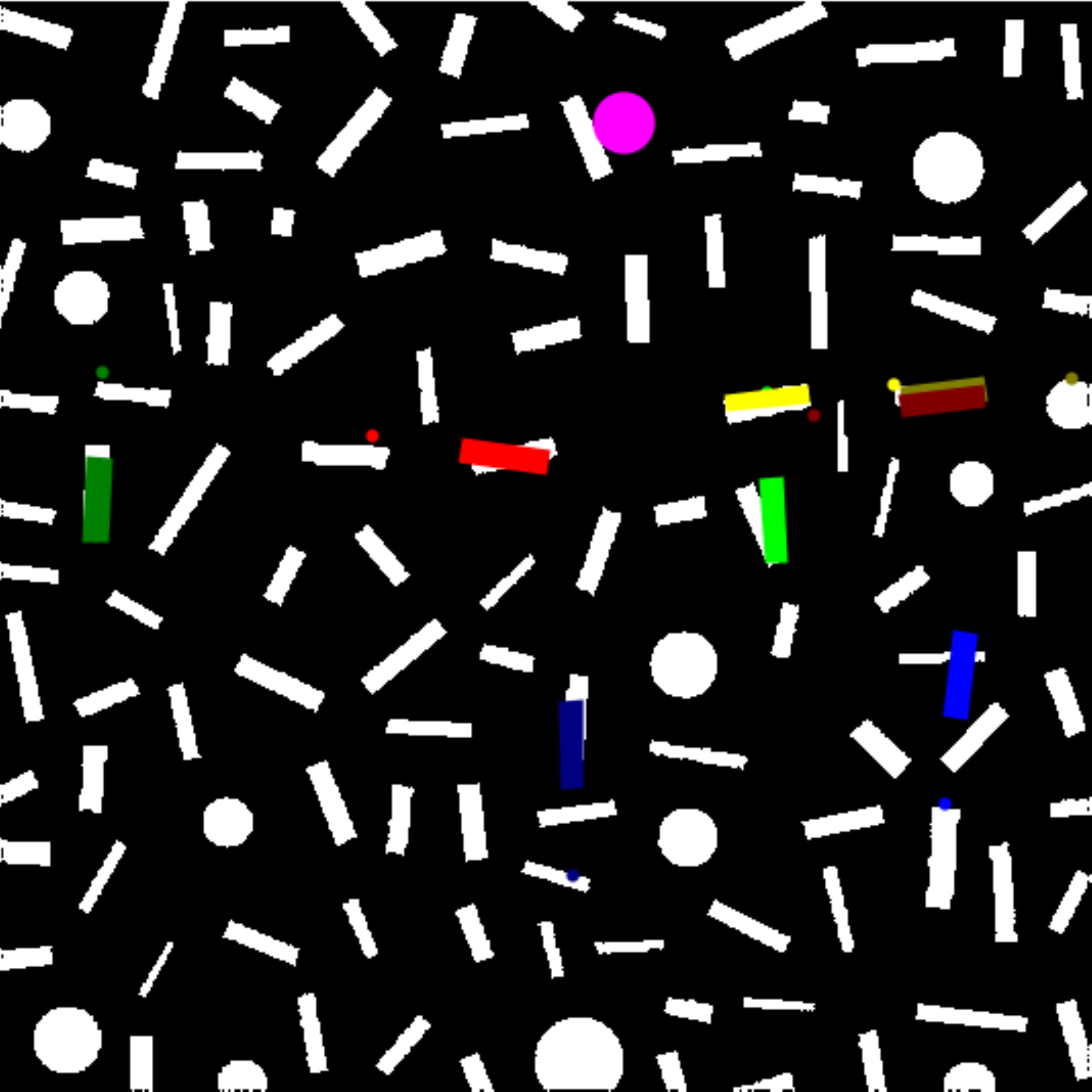}} ~~
\subfloat[MLE at iteration 10]{\includegraphics[width=0.24\columnwidth]{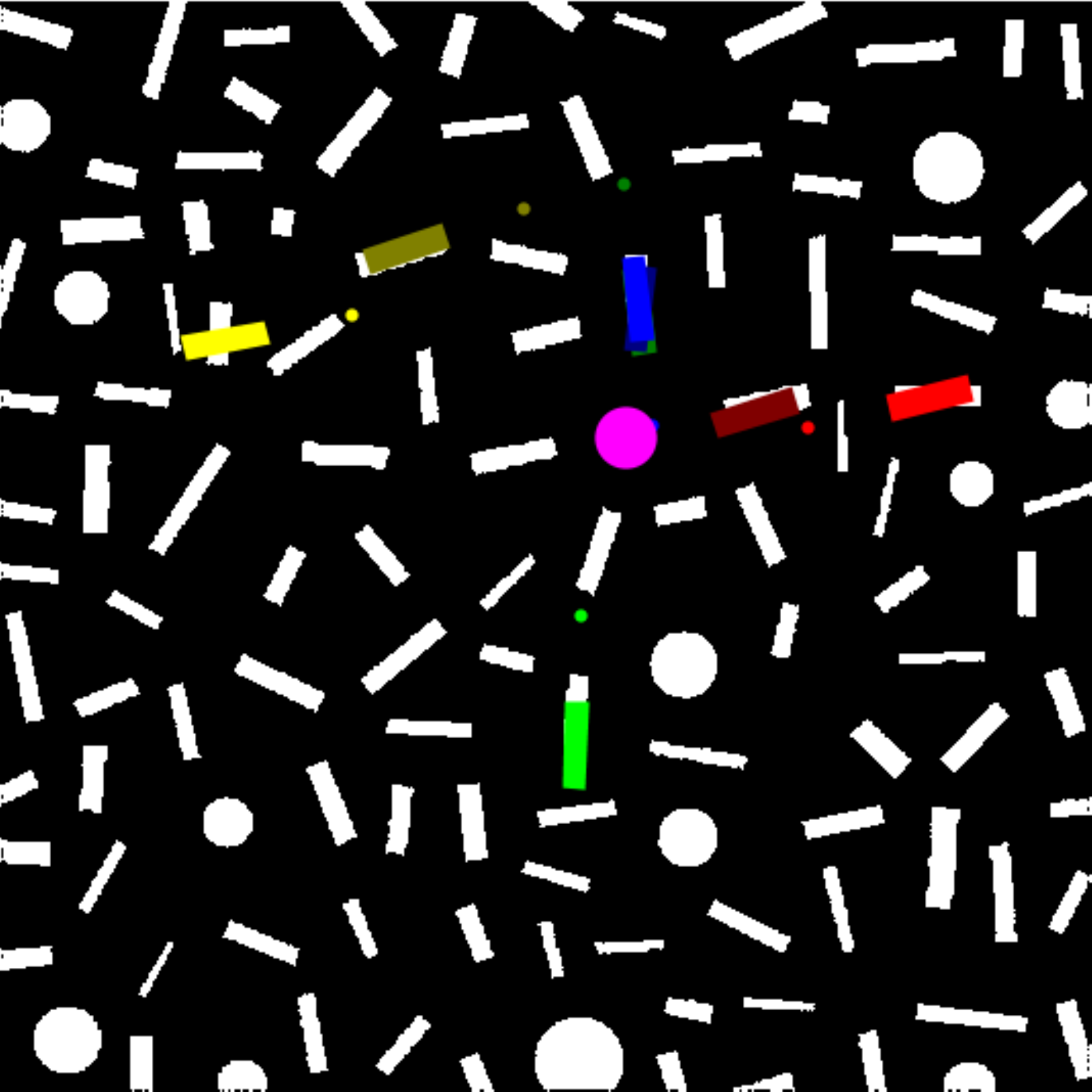}} ~~
\subfloat[MLE at iteration 34]{\includegraphics[width=0.24\columnwidth]{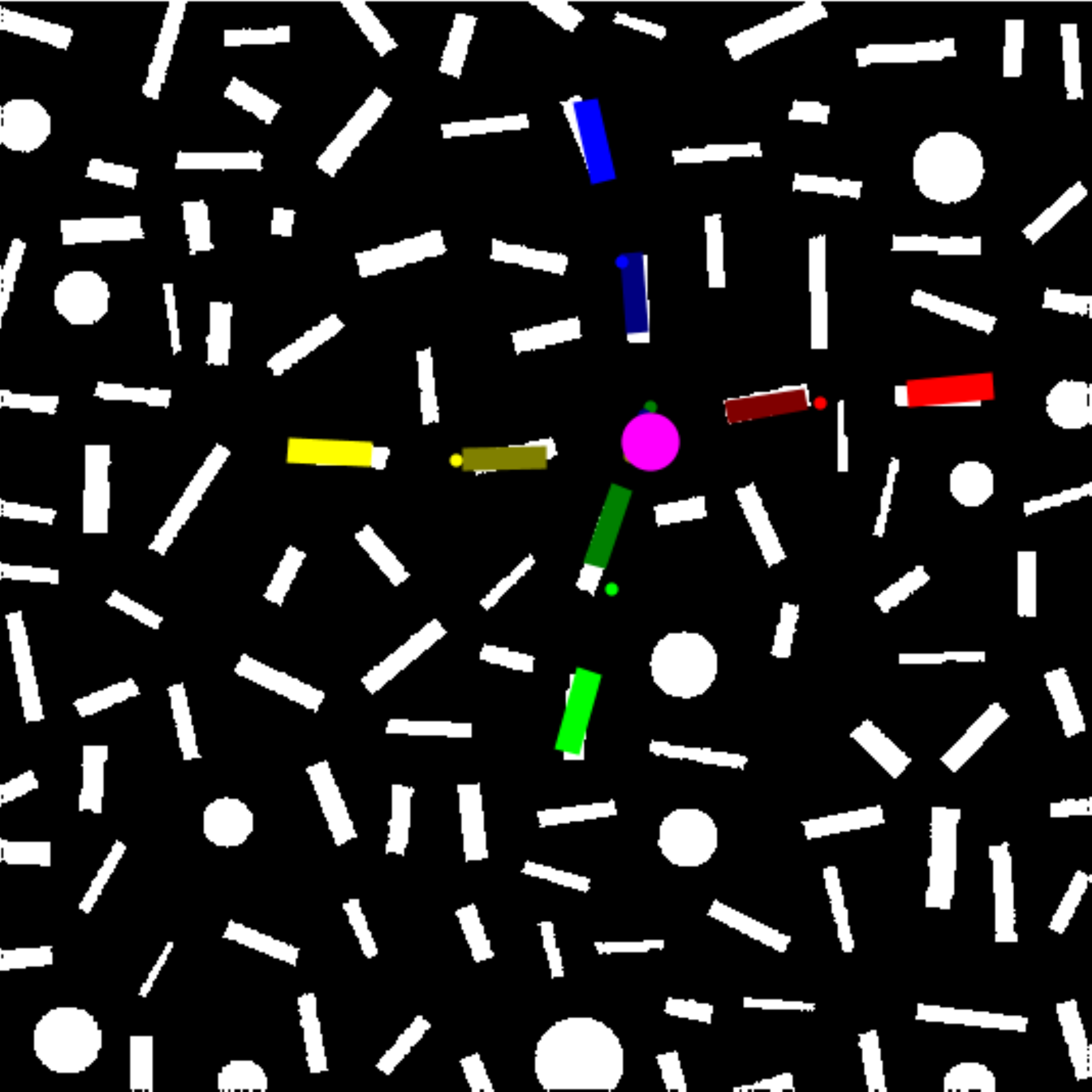}} \\
 \caption{\footnotesize PMPNBP results with circle node ``occluded''. Each message contains 200 particles initialized randomly at locations where their $\phi_s > 0.4$. The top row shows the belief samples $bel_s$ for each of the nodes and the bottom row shows their Maximum Likely Estimate (MLE). MLE at iteration 34 has all the links and circle converged to their ground truth states (Best viewed in color).}
 \label{pmpnbp_results_no_circle}
\end{figure}

We show the convergence of the PMPNBP qualitatively in Figure~\ref{pmpnbp_results_circle} and Figure~\ref{pmpnbp_results_no_circle}. The pattern referred in Figure~\ref{link_structure} is placed in a clutter made of 12 circles and 100 rectangles. There are 16 messages, i.e., 4 from circle to inner links, 4 from inner links to circle, 4 from inner links to outer links and 4 from outer to inner links. The initialization of the messages is done with $N=75$ particles at $(x, y)$ locations of the image where $\phi_s > 0.4$. This is assumed to be the coarse feature detection of the circle and rectangles in the image replicating the initialization in ~\cite{isard2003pampas}. In the future iterations, the message $m_{t\rightarrow s}$ has $50\%$ of the samples uniformly sampled in the image to keep exploring, while the other $50\%$ of the samples are sampled from the marginal belief $bel_s$. As it can be seen in Figure~\ref{pmpnbp_results_circle}, the initialization (Belief at Iteration 0) is distributed across the image. At iteration 1, the message passing starts to influence the belief of the nodes and at iteration 10, they form the spatial arrangement satisfying their geometrical structure. At iteration 24, the most likely estimate of all the links and circle are close to the pose of the ground truth pattern. 

The second example in Figure~\ref{pmpnbp_results_no_circle} has no circle in the center of the pattern, demonstrating an occlusion scenario. This scenario demonstrates that the proposed algorithm retains the power of the probabilistic modeling of the belief propagation approach. The initialization is done similar to the first example, where there were no samples near the "occluded" circle. The convergence is similar to the first example but takes 34 iterations to converge.

\begin{figure}[t!]
    \centering
    \captionsetup[subfigure]{labelformat=empty}
\subfloat[(a) Average error vs Iterations]{\includegraphics[width=0.5\columnwidth]{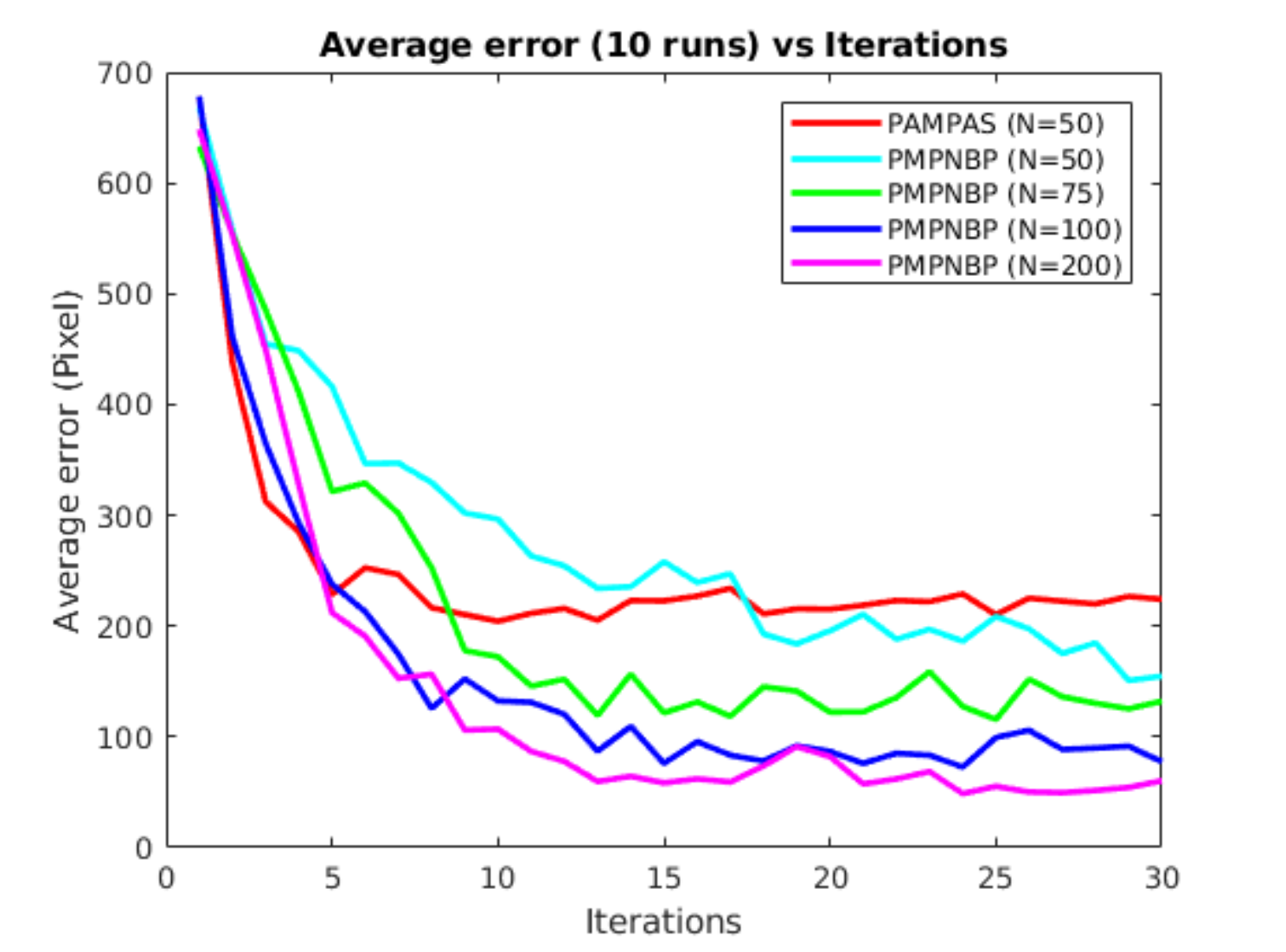}}
\subfloat[(b) CPU time vs Particles]{\includegraphics[width=0.5\columnwidth]{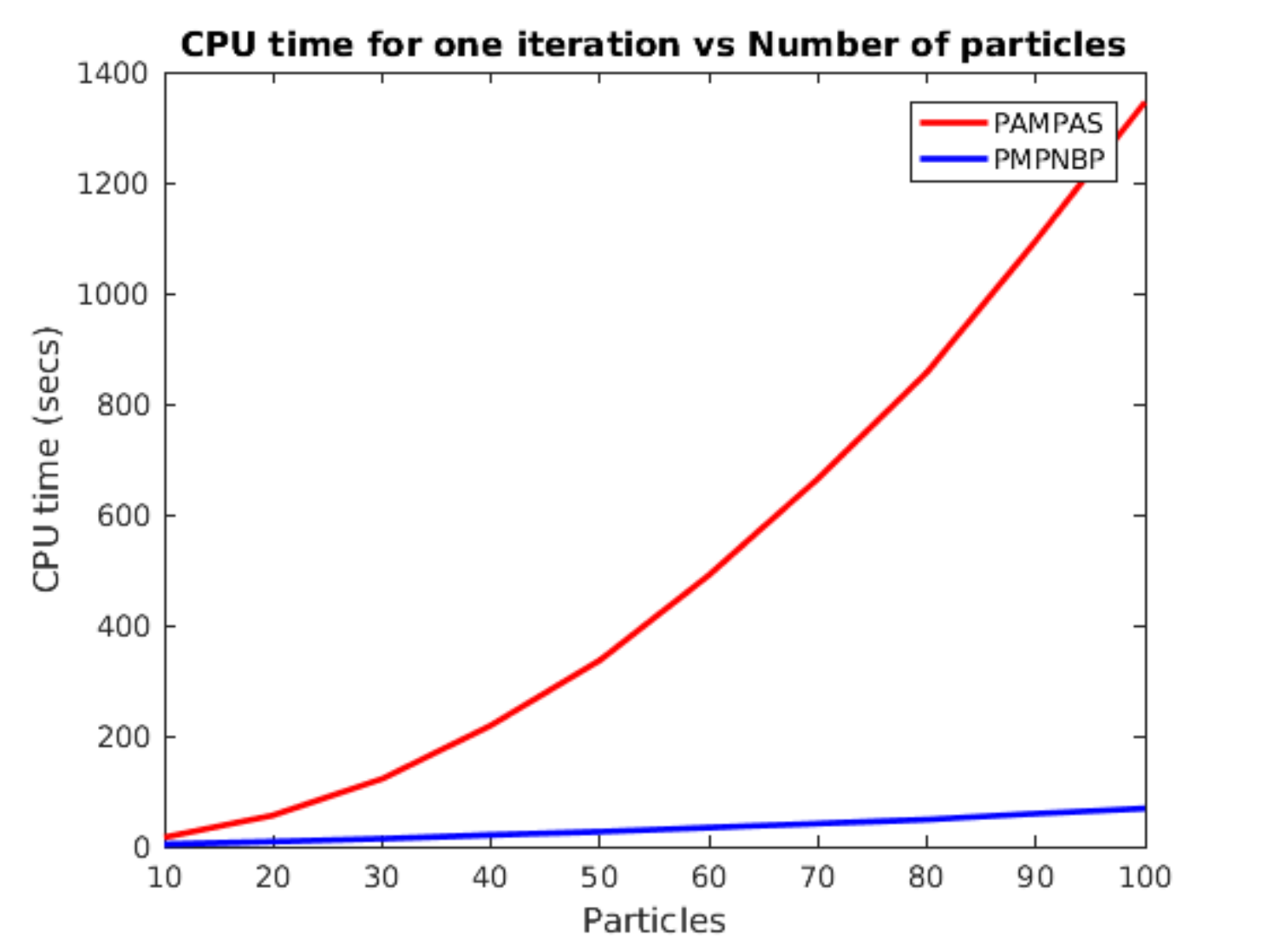}} 
 \caption{\footnotesize Convergence and execution time: (a) shows the average position error of Maximum Likely Estimate (MLE) achieved by PMPNBP ($N=\{50, 75, 100, 200\}$) in comparison to PAMPAS (our implementation) for the  experiment in Figure~\ref{pmpnbp_results_circle}. (b) shows CPU time per iteration required for PMPNBP and PAMPAS, as the number of particles grow. This shows the PMPNBP achieves comparable convergence with efficient computation.}
 \label{plots}
\end{figure}

In Figure~\ref{plots}(a), we show the convergence of the PMPNBP with respect to the previous algorithm PAMPAS ~\cite{isard2003pampas} which uses Gaussian mixture to represent the messages and use Gibbs sampler to perform message products (for circle). Convergence here is shown as the average error of the Maximum Likely Estimate from its ground truth with respect to the number of belief iterations over 10 trials. We plot this convergence for PMPNBP with $N=\{50, 75, 100, 200\}$ components versus PAMPAS with $N=50$. The convergence of PMPNBP is better than our implementation of PAMPAS. It can also be noted that the PMPNBP has decreasing average error with increasing numbers of particles. This essentially indicates that as larger $N$ the better the inference will be. To evaluate whether PMPNBP accommodates the use of larger $N$ in practice, we plot the CPU run time per message update iteration in Figure~\ref{plots}(b). An entire message generation in PAMPAS takes $O(KDN^2)$ operations, where $D$ is the number of messages to compute product in the ``pre-message'', $K$ is the number of iterations for the Gibbs sampler and $N$ is the number of components representing a message. In contrast, PMPNBP takes only $O(N)$ operations. For the plots in Figure~\ref{plots}(b) with PAMPAS we use $K=50$ as the Gibbs sampler iterations. 

These results indicate that the proposed PMPNBP has similar convergence properties as the earlier approaches with greater computational efficiency.

\section{Conclusion}\label{conclusion}
We proposed a new message passing scheme that uses a ``pull'' approach to update messages in Nonparametric Belief Propagation. We represent messages as weighted particles instead of Gaussian Mixtures as proposed in earlier algorithms. The proposed message passing scheme avoids Gibbs sampling based message products of the earlier methods and provides faster product approximations. We show the efficiency of the proposed algorithm both in terms of its convergence properties and the computing time with respect to PAMPAS. The 2D illustration chosen in this paper, is suggestive of the real-world problems in scene estimation. We believe that the results are promising enough to stimulate further research in applying PMPNBP algorithm for scene estimation problems, especially in robotic perception where a notion of uncertainty in the inference is inevitable.


\section*{Appendix A}
\label{appendix_a}
The details of pairwise sampling procedure described in Section~\ref{experimental_setup} are provided here. $X_1$, $X_i$ and $X_j$ denote the circle node, inner link nodes and outer link nodes respectively, where $2 \leq i \leq 5$ and $j=i+4$.
The samples generated for outer arms $j=i+4$ given an inner arm $2\leq i \leq 5$ using
\begin{gather*}
x_j = x_i + w_i \cos(\alpha_i) + \mathcal{N}(.;0, \sigma_p^2) \\
y_j = y_i + w_i \sin(\alpha_i) + \mathcal{N}(.;0, \sigma_p^2) \\
\alpha_j = \alpha_i + \mathcal{N}(. ; 0, \sigma_{\alpha}^2) \\
w_j = w_i +\mathcal{N}(.; 0, \sigma_s^2) \\
h_j = h_i +\mathcal{N}(.; 0, \sigma_s^2)
\end{gather*}

The samples for inner arms $2\leq i \leq 5$ are generated given an outer arm $j=i+4$ using
\begin{gather*}
\hat{x}_i = x_j+\mathcal{N}(.;0, \sigma_p^2) \\
\hat{y}_i = y_j+\mathcal{N}(.;0, \sigma_p^2) \\
\hat{\alpha}_i = \pi+\alpha_j \\
x_i = \hat{x}_i + w_j \cos(\hat{\alpha}_i) + \mathcal{N}(.;0, \sigma_p^2) \\
y_i = \hat{y}_i + w_j \sin(\hat{\alpha}_i) + \mathcal{N}(.;0, \sigma_p^2) \\
\alpha_i = \hat{\alpha}_i -\pi+ \mathcal{N}(. ; 0, \sigma_{\alpha}^2)\\
w_i = w_j +\mathcal{N}(.; 0, \sigma_s^2) \\
h_i = h_j +\mathcal{N}(.; 0, \sigma_s^2)
\end{gather*}


The samples for circle are generated given an inner arm $2 \leq i \leq 5$ using
\begin{gather*}
x_1 = x_i + \mathcal{N}(.;0, \sigma_p^2) \\
y_1 = y_i + \mathcal{N}(.;0, \sigma_p^2) \\
r_1 = 0.5(w_i/\delta_w + h_i/\delta_h)+ \mathcal{N}(.; 0, \sigma_s^2)
\end{gather*}

The samples for inner arms $2 \leq i \leq 5$ are generated given the circle using
\begin{gather*}
x_i = x_1 + \mathcal{N}(.;0, \sigma_p^2) \\
y_i = y_1 + \mathcal{N}(.;0, \sigma_p^2) \\
\alpha_i = i\times\pi/2 + \mathcal{N}(. ; 0, \sigma_{\alpha}^2)\\
w_i = \frac{2Cr_1\delta_w\delta_h}{C\delta_h+\delta_w} + \mathcal{N}(.; 0, \sigma_s^2) \\
h_i = \frac{2r_1\delta_w\delta_h}{C\delta_h+\delta_w} + \mathcal{N}(.; 0, \sigma_s^2)
\end{gather*}

The parameters for the dimensions are $C=7$, $\delta_w=\frac{28}{5}$ and $\delta_w=\frac{4}{5}$. The values used as the standard deviations are $\sigma_p=10$, $\sigma_s=2$, $\sigma_\alpha=15\degree$. These values are set constant for all the results in this paper.


\end{document}